\documentclass[conference]{IEEEtran}
\IEEEoverridecommandlockouts
\usepackage{cite}
\usepackage{amsmath,amssymb,amsfonts}
\usepackage{graphicx}
\usepackage{textcomp}
\usepackage{xcolor}
\usepackage{algorithm}

\usepackage[noend]{algpseudocode}
\newcommand*{\defeq}{\stackrel{\text{def}}{=}}
\usepackage{amsthm}
\usepackage{subcaption}
\newcommand{\ve}[1]{{\mbox{\boldmath${#1}$}}}
\theoremstyle{definition}
\newtheorem{definition}{Definition}
\usepackage{balance}
\usepackage[bookmarks=false]{hyperref}

\def\BibTeX{{\rm B\kern-.05em{\sc i\kern-.025em b}\kern-.08em
    T\kern-.1667em\lower.7ex\hbox{E}\kern-.125emX}}
\begin{document}

\title{Regularized Operating Envelope with Interpretability and Implementability Constraints}

\author{\IEEEauthorblockN{Qiyao Wang, Haiyan Wang, Chetan Gupta, Susumu Serita}
\IEEEauthorblockA{Industrial AI Lab, Hitachi America, Ltd. R$\&$D \\
Santa Clara, CA \\
firstname.lastname@hal.hitachi.com}
}

\maketitle
\footnotetext[1]{978-1-7281-0858-2/19/\$31.00 \copyright 2019 IEEE. Personal use of this material is permitted.  Permission from IEEE must be obtained for all other uses, in any current or future media, including reprinting/republishing this material for advertising or promotional purposes, creating new collective works, for resale or redistribution to servers or lists, or reuse of any copyrighted component of this work in other works.}

\begin{abstract}
Operating envelope is an important concept in industrial operations. Accurate identification for operating envelope can be extremely beneficial to stakeholders as it provides a set of operational parameters that optimizes some key performance indicators (KPI) such as product quality, operational safety, equipment efficiency, environmental impact, etc. Given the importance, data-driven approaches for computing the operating envelope  are gaining popularity. These approaches typically use classifiers such as support vector machines, to set the operating envelope by learning the boundary in the operational parameter spaces between the manually assigned `large KPI' and `small KPI' groups. One challenge to these approaches is that the assignment to these groups is often ad-hoc and hence arbitrary. However, a bigger challenge with these approaches is that they don't take into account two key features that are needed to operationalize operating envelopes: (i) interpretability of the envelope by the operator and (ii) implementability of the envelope from a practical standpoint. In this work, we propose a new definition for operating envelope which directly targets the expected magnitude of KPI (i.e., no need to arbitrarily bin the data instances into groups) and accounts for the interpretability and the implementability. We then propose a regularized `GA + penalty' algorithm that outputs an envelope where the user can tradeoff between bias and variance. The validity of our proposed algorithm is demonstrated by two sets of simulation studies and an application to a real-world challenge in the mining processes of a flotation plant. 
\end{abstract}


\section{Introduction}
In industrial operations, an important concept is that of the operating envelope. Conceptually, the operating envelope is a set of operational parameters, such that some KPI is optimized. In the industrial context, typical KPIs include product quality, operational safety, equipment efficiency, environmental impact, etc \cite{wang2019multilayer, zhang2018equipment, wang2018maintenance, wang2019remaining}. The operating envelope has wide application since it directly targets the business outcome and yields actionable recommendations in the operations space. It has been widely considered across numerous industrial domains. For instance, an operating envelope for an oil well was identified by \cite{kassim2013holistic} to ensure asset integrity and production performance in the oil and gas field. Similarly, an operating range in terms of airflow, fuel cell, and turbine load was determined to ensure the steadiness of a direct fired fuel cell gas turbine \cite{tucker2009determination}. Other published applications of the operating envelope include generating a safer industrial system design (e.g., sub-sea oil production facilities) \cite{stuber2013evaluation}, and identifying more stable operations for partially stable combustion engines \cite{janakiraman2013modeling}. 

Despite its popularity and significance in the industrial domain, there is no universally agreed upon definition for an operating envelope. Very often, different concepts are ambiguously referred to as an operating envelope by different groups of researchers and engineers. The first contribution of our paper is that we comprehensively identify the required components to define the operating envelope and systematically distinguish among different types of operating envelopes. 

Analogous to traditional machine learning, the target value or the KPI of interest is called the \textit{response variable}. Without loss of generality, we assume that there is only one response variable of interest and higher response values indicate better performances. The factors that affect the response variable can be categorized into either \textit{control variables} or \textit{state variables}. The \textit{control variables} correspond to operations that are fully controllable by operators. From modeling perspective, these variables can be treated as deterministic. For instance, when driving a car, the pressure from feet to the pedal is a control variable. The \textit{state variables} quantify factors which affect the response variable but cannot be directly controlled. State variables are considered as random variables in modeling. In the car driving example, the actual pressure transmitted to the accelerator is a state variable which affects the response variable speed, but cannot be directly controlled.

Depending on the availability of data sources and the problem of interest, three types of operating envelopes and the corresponding use cases can be defined, i.e., 1) operating envelopes with respect to the state variables, 2) operating envelopes with respect to control variables, 3) operating envelope with respect to both state and control variables. In this paper, we mainly focus on the first type of operating envelope which corresponds to scenarios where the response variable and state variables are given, while the control variables are not recorded or not easily quantifiable. The impact of control variables on the response variable is through the state variables, and the underlying mechanics are typically not completely understood or too difficult to model. The operating envelope problem is then to identify good sets of values in the state variable space such that the corresponding response variable is higher or highest when compared to the other regions. Random state variables are the more general cases and more challenging to be dealt with than the deterministic control variables. The proposed definition and the corresponding solutions discussed later can easily be extended to handle the other two types of scenarios with appropriate modifications.

In the literature, for the considered type of operating envelope problem, there are two existing types of methods. The first type of methods is based on domain knowledge. Specifically, domain experts build physical models to determine or simulate the specific optimal operating envelope. Previous work falling into this category include \cite{tucker2009determination, rallo1986operating}. This type of approach is well supported by the domain expert's opinions. However, this manual operating envelope identification process is time-consuming and even infeasible when the industrial process or equipment is so complicated that there are too many correlated state variables that need to be considered simultaneously. Another drawback of domain knowledge based approaches is that they cannot be easily scaled across different industries or different types of equipment.

The second type of approaches uses data-driven methods. For these, data records are first grouped into a `high response' and a `low response' groups based on the response variable. Then classification algorithms are utilized to learn the boundary between these two groups. The learned boundary is used to indicate the operating envelope, i.e., regions in the state variable space with a higher probability of belonging to the `high response' group are considered acceptable. For example, in \cite{kolmanovsky2002support} authors used the support vector machine to determine the gasoline direct injected engine admissible operating envelope. Classification based approaches have several drawbacks: (i) When the response variable is a numerical variable, classification based algorithms discard information about the response variable by converting the numerical data to categorical data. The lack of differentiation on the KPI values within the identified region may result in sub-optimal performance.
(ii) A bigger challenge with these approaches is that with the goal of maximizing classification accuracy, these approaches may result in very complex or narrow regions, which greatly reduces the \textit{interpretability} (meaning, whether the outcome is comprehensible by the operator) and \textit{implementability} (meaning that the operating envelope can be achieved in practice) of the operating envelope. For example, the classification algorithms usually do not put continuity constraints on the targeting operating envelope. As an illustration, suppose we have three state variables denoted by $a$, $b$, $c$, and we apply classification algorithms to identify the operating envelope in the space spanned by these three state variables to achieve `high' performance. The resulting operating envelope with high classification accuracy may indicate that states $(a_0,b_0,c_0)$  and $(a_0+0.2,b_0,c_0)$ are good operations and are inside the envelope, but the another state in-between the line connecting the above two states, $(a_0+0.05,b_0,c_0)$, is a bad operation and is outside of the envelope. This type of non-intuitive operating envelope is not comprehensible to operators and is an obstacle in operationalizing it in practice. That is, the operating envelopes of complex shapes that are \textit{non-compact} are not easily interpretable. Another observation is that tight and narrow operating envelopes are not easily implementable. 

To overcome these limitations, we propose a new definition for the operating envelope problem. Aiming at solving the first limitation listed above, we propose to directly set the expected value of the response variable as the optimization target. As for the second set of limitation, to ensure interpretability, we only consider operating envelopes that are in the form of a single parameterizable continuous compact set or union of such disjoint compact sets in the state variable space, and to ensure the implementability of the resulted operating envelope, we put an additional constraint on the probabilistic coverage of our recommendation. Specifically, we require that the probability of state variables falling into the operating envelope is greater than some pre-specified threshold. We call the identification of the region with the largest mean response value subjecting to these interpretability and implementability constraints as the \textit{operating envelope with interpretability and implementability constraints problem}. (We give precise definitions in the next section). This is a novel way of formulating the operating envelope problem, which is motivated by the fact that real-world practitioners usually not only care about the accuracy of our recommendation but also the interpretability and the implementability of data-driven solutions.


Interpretable machine learning is a rapidly rising field \cite{lundberg2017unified, gilpin2018explaining} with the growing implications of deep learning models in real world practices. The classical interpretable models focus on interpreting the impact of certain feature values on the model predictions. We extend this idea further: the interpretability in our work aims to ensure that the recommendation on the identified variables can be easily interpreted by operators. For example, the value of a certain variable should be in a range. Like the traditional understanding of interpretability, we aim at increasing the trust between human and models, and hence their acceptance.

Our proposed operating envelope problem is essentially a region searching task. The identifications of interesting regions has been previously studied by, such as the detection of outlying regions \cite{liu2008isolation}, the detection of salient regions in computer vision \cite{cheng2014global}, the detection of clusters, etc. However, none of these region identification problems is formulated to maximize the expected region-wise response variable value. The proposed operating envelope with interpretability and implementability constraints has never been considered before and no existing algorithms can be directly applied to solve it. In this paper, we propose an effective solution which is summarized as follows.

We first apply the penalty method \cite{bertsekas2014constrained} to turn the probabilistic coverage constrained optimization problem into an unconstrained problem. As shown in the next section, no closed form exists for the objective function in general, which eliminates the applicability of methods that require gradient or Hessian information (e.g., Newton's method). One may then adapt one-dimensional directed search methods such as Golden Section Search to solve this multi-dimensional  optimization problem with black-box function via coordinate cycling search. However, the directed search based methods process a single point of the search space in the searching iterations and are not population based, which poses a high risk of being trapped in a local optimum. Hence, we propose to use the genetic algorithm (GA) \cite{chelouah2000continuous} to search over all possible regions characterized by a finite set of parameters. GAs are more robust than the directed search algorithms and the effectiveness depends less on the initial solution values. Another important property of the GAs is that they maintain a population of potential solutions, which reduces the risk of being trapped in a local minimum. 
To boost the efficiency of GA, we adopt parallelization for GA \cite{muhlenbein1991parallel}. To enhance the generalization of the calculated operating envelope, we propose to add a regularization term using the sample standard deviation of the estimated expected response value within any region during the learning phase. 

The contributions of the paper are summarized as follows:
\begin{itemize}
\item We systematically identify the components in the conventional operating envelope problem and map the space of  existing solutions.
\item We propose the interpretability and the implementability constraints, and define a new operating envelope identification problem with these two constraints. 
\item For the new operating envelope definition, we design an algorithm by combining the penalty method with the genetic algorithm. 
\item We also propose a regularization strategy to improve the generalization of the achieved operating envelope.
\end{itemize}

The rest of the paper is organized as follows. Section \ref{sec2} presents the proposed definition for the \textit{operating envelope with interpretability and implementability constraints problem}. Section \ref{sec3} discusses in detail our proposed solutions, including the estimation module, the search module, and the generalization consideration. Section \ref{sec4} presents numerical experiments including two simulation studies and one real-world data analysis task. Section \ref{sec5} concludes the paper.

\section{Definition of the Proposed Operating Envelope with Interpretability and Implementability Constraints Problem}\label{sec2}

\subsection{Mathematical notations}

Let $Y$ denote the response variable, which represents the KPI of interest in our operating envelope problem. The state variables that affect the value of $Y$, i.e., the \textit{input} in machine learning language, are denoted by a $p-$dimensional vector $\mathbf{X}=[X_1, ...,X_j,..., X_p]^T$.  Denote the underlying density distribution of $\mathbf{X}$ as $g(\mathbf{X})$, which is unknown. To simplify our notations, we mainly focus on scenarios where the $p$ state variables are numerical variables in this paper. The proposed problem formulation and the corresponding solutions can be extended to handle categorical state variables with mild modifications. The space spanned by $\mathbf{X}$ is denoted as $\Omega(\mathbf{X})$. Suppose that there is an underlying mapping from $\mathbf{X}$ to the response variable $Y$, i.e., 
\begin{equation}
\label{ge_eq1}
Y=f(\mathbf{X}) + \epsilon, 
\end{equation}
where $\epsilon$ represents the combination of any other factors that is not in $\mathbf{X}$ and the random disturbance. We assume that the mapping has been normalized such that the expectation of $\epsilon$ is 0 and the variance is $\sigma_{\epsilon}^2$, and $\epsilon$ is independent of $\mathbf{X}$. Neither the structure of $f(\cdot)$ nor $\sigma_{\epsilon}^2$ is known beforehand. Instead, we have access to $n$ independent pairs of samples, $(y_i, \mathbf{x}_i)$, with $\mathbf{x}_i=[x_{1}^{(i)}, ...,x_{j}^{(i)},..., x_{p}^{(i)}]^T$ for $i=1,...,n$.

\subsection{Problem formulation}

The ultimate goal of our operating envelope problem is to identify an optimal region in the state variable space $\Omega(\mathbf{X})$ such that the response variable is maximized, and the detected region is interpretable and implementable by making use of the observed data points $(y_i, \mathbf{x}_i)$, $i=1,...,n$. Before presenting the proposed formulation for our new operating envelope problem, we first describe the concepts of interpretability and implementability of any region $R$ in the state variable space.

To enhance the interpretability of the achieved envelope, we propose to define eligible candidates as regions of certain simple geometrical shapes. Specifically, we assume that the qualified regions are $p$-sided hyper-rectangles with sides being parallel to the $p$ state variable axes or unions of $L$ ($L<\infty$) disjoint such hyper-rectangles. The mathematical representations of this type of hyper-rectangles is provided in Definition \ref{def}. Hyper-rectangles that are parallel to the orthogonal state variable axes are essentially the Cartesian product of intervals on each axis. The identified optimal hyper-rectangle explicitly indicates within which interval each state variable should be, increasing the explainability of data-driven results to operators. 

For any region $R$ defined above, we propose to use the probability that state variable $\mathbf{X}$ falling into it to quantify its implementability, i.e. $P_R=P(\mathbf{X}\in R)$. A larger value of $P_R$ indicates that the corresponding region has a better implementability. To ensure the implementability in the output region, we introduce a lower bound $\beta\in[0,1)$ and define implementable regions as all regions that satisfy the coverage constraint
\begin{equation}
\label{eq2}
P_R=P(\mathbf{X}\in R) > \beta.
\end{equation}
In Eq.~\eqref{eq2}, higher values of $\beta$ place higher requirements on the implementability. If any region size is implementable, $\beta$ can be set as 0. $\beta$ is a hyperparameter that needs to be specified in advance according to the practical needs.

Based on the described interpretability and implementability concepts, the proposed operating envelope problem is formally defined in Definition \ref{def} below.

\theoremstyle{definition}
\begin{definition}\label{def}
For a response variable $Y$ and $p$ state variables $\mathbf{X}=[X_1, ...,X_j,..., X_p]^T$, suppose that any targeting operating envelope is an union of $L$ ($0<L<\infty$) disjoint hyper-rectangles in the state variable space $\Omega(\mathbf{X})$, which is denoted by $R=R_1 \bigcup...R_l...\bigcup R_L$, where $R_l=[r_{l1}^{l}, r_{l1}^{u}]\times...\times[r_{lj}^{l}, r_{lj}^{u}]\times...\times[r_{lp}^{l}, r_{lp}^{u}]$, with $r_{lj}^{l}, r_{lj}^{u}\in \mathbb{R}$ and $-\infty<r_{lj}^{l}< r_{lj}^{u}<\infty$, for $l=1,...,L$ and $j=1,...,p$.

The operating envelope is then the solution of:
\begin{equation}
\label{def_eq1}
    \max_{R} E[Y|\mathbf{X}\in R]
\end{equation}
subject to the probabilistic coverage constraint
\begin{equation}
\label{def_eq2}
P_R=P(\mathbf{X}\in R) > \beta, \text{with} \,\,\beta\in[0,1),
\end{equation}
as well as the mutually disjoint region constraint, i.e., $\forall l, l^\prime \in \{1,...,L\}$ and $l \neq l^\prime$, there exist $j\in \{1,...,p\}$ such that 
\begin{equation}
\label{def_eq3}
\max({r_{lj}^{l}, r_{l^\prime j}^{l}}) > \min({r_{lj}^{u}, r_{l^\prime j}^{u}}).
\end{equation}
\end{definition}

\begin{figure}[htbp]
	\centering
	\includegraphics[width=50mm]{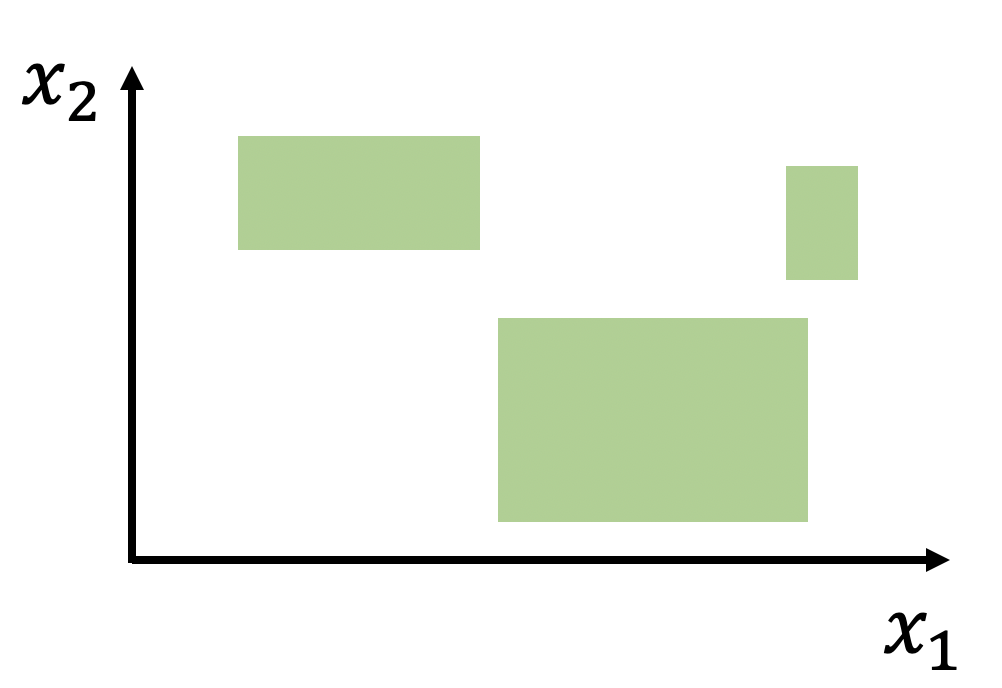} 
	\caption{Visualization of the desired operating envelope with $L=3$ and $p=2$ in Definition \ref{def}.}\label{Fig:def}
\end{figure}

In Definition \ref{def}, Eq.~\eqref{def_eq1}, the main objective function, indicates that the goal of our proposed operating envelope is to maximizing the region-wise response variable over $2pL$ region-characterizing factors, i.e, $r_{l1}^{l}, r_{l1}^{u},...,r_{lp}^{l}, r_{lp}^{u}$, for $l=1,...,L$. To the best of our knowledge, this is the first formal definition for the operating envelope identification problem that directly considers the expected value of response variable $Y$ within regions as the objective function. In previous literature on data-driven operating envelope problems, machine learning classifiers treat the accuracy measures for predicting the labels `$Y \geq$  cutoff' and `$Y <$  cutoff'\cite{shrikumar2017learning, gunning2017explainable} as the objective function. The proposed definition also originally incorporates customizable interpretability and implementability constraints by Eq.~\eqref{def_eq2} and Eq.~\eqref{def_eq3} to account for important practical requirements. The desired operating envelope that is a union of $L=3$ disjoint rectangles within a space spanned by $p=2$ covariates is given in Figure \ref{Fig:def}.

Note that our proposed definition can be generalized to handle candidate regions with other shapes that can be characterized by a finite set of parameters, such as hyperspheres. In Definition \ref{def}, $L$ is a hyperparameter that indicates the number of desired disjoint hyper-rectangles in the output. Specifying multiple disjoint sub-regions in the output is sometimes important. For instance, the operations may be conducted at multiple different operating modes. For each mode, we need to specify an optimal operating region accordingly. This hyperparameter is optional for our constrained operating envelope problem. When it is not specified, we can output the operating envelope and the corresponding optimal mean response variable under different $L$, and the user can choose among the results based on their own needs. 

In the next section, we describe step by step how we use the sample data $(y_i, \mathbf{x}_i)$, $i=1,...,n$, to obtain the optimal envelope defined in Definition \ref{def}.

\section{A Data-Driven Solution Approach for the New Operating Envelope Problem}\label{sec3}
To solve the problem defined in the previous section, two essential modules are required. First, an estimation module is needed to estimate the objective function $E[Y|\mathbf{X}\in R]$ and the probabilistic coverage $P_R=P(\mathbf{X}\in R)$ for any given region $R$. Proposed estimation methods are described in Section \ref{sec3.1}. Second, a search module should be constructed to globally search over the constrained state variable space. Relevant proposals are provided in Section \ref{sec3.3} and Section \ref{sec3.4}. An operating envelope specific generalization problem and the proposed regularization technique are presented in Section \ref{sec3.2}. 

\subsection{Estimate objective function and probabilistic coverage} \label{sec3.1}

%

A natural way of evaluating the objective function as well as the probabilistic coverage for any region $R$ could be that we first estimate $f(\mathbf{X})$ by $\hat{f}(\mathbf{X})$, for example, through regression models, and estimate $g(\mathbf{X})$, for example, via the kernel density estimation $\hat{g}(\mathbf{X})$. Then we estimate the objective function and the probabilistic coverage by plugging in $\hat{f}(\mathbf{X})$ and $\hat{g}(\mathbf{X})$ to the objective function and the probabilistic coverage, respectively. Specifically, the estimates are
\begin{equation}
\label{est_eq11}
\tilde{E}[Y|\mathbf{X}\in R]=\frac{\hat{\int_R}\hat{f}(\mathbf{X})\hat{g}(\mathbf{X})d\mathbf{X}}{\hat{\int_R}\hat{g}(\mathbf{X})d\mathbf{X}},
\end{equation}
and
\begin{equation}
\label{est_eq22}
\tilde{P}_R=\hat{\int_R}\hat{g}(\mathbf{X})d\mathbf{X}, 
\end{equation}
where $\hat{\int_R}$ represents the numerical integration using techniques reviewed by \cite{davis2007methods}. However, this function approximation first and then numerical integration method is computationally heavy, especially in the high dimensional data cases.

To overcome the calculation challenge, we proposed an alternative model-free approach. Based on the observed samples $(y_i, \mathbf{x}_i)$, $i=1,...,n$, for any given region $R$, we propose to use the sample average approximation (\textit{SAA}) to estimate $E[Y|\mathbf{X}\in R]$. Let $I(U)$ be an indicator function of event $U$, i.e., 
\begin{equation}
    I(U) = \left \{
  \begin{aligned}
    &0, && \text{Event } U\text{   is False} \\
    &1, && \text{Event } U\text{   is True}
  \end{aligned} \right., 
\end{equation}
then the proposed estimator for $E[Y|\mathbf{X}\in R]$ is 
\begin{equation}
\label{est_eq1}
\hat{E}[Y|\mathbf{X}\in R]=\frac{\sum_{i=1}^n y_i I(\mathbf{x}_i\in R)}{\sum_{i=1}^n I(\mathbf{x}_i\in R)}.
\end{equation}
For any given region $R$, we propose to estimate the probabilistic coverage $P_R=P(\mathbf{X}\in R)$ by the sample proportion estimator. Mathematically, 
\begin{equation}
\label{est_eq2}
\hat{P}_R=\frac{\sum_{i=1}^n I(\mathbf{x}_i\in R)}{n}.
\end{equation}

%

The model-free sample average estimator and the sample proportion estimator in Eq.~\eqref{est_eq1} and Eq.~\eqref{est_eq2} are efficient to calculate. Our simulation studies also indicate that they yield accurate estimates for $E[Y|\mathbf{X}\in R]$ and $P_R=P(\mathbf{X}\in R)$, respectively, as long as the sample data is representative over the entire space $\Omega(\mathbf{X})$. 

When the raw data  $(y_i, \mathbf{x}_i)$, $i=1,...,n$ is not representative over the entire space. We propose to improve data's representability by first generating more data from the learned models $\hat{f}(\mathbf{X})$ and $\hat{g}(\mathbf{X})$ before using  Eq.~\eqref{est_eq1} and Eq.~\eqref{est_eq2} to estimate the required components. We focus on the model-free estimation approach given in Eq.~\eqref{est_eq1} and Eq.~\eqref{est_eq2} in this paper.  



\subsection{Penalty method for the constraints}\label{sec3.3}
Based on the estimation module in the previous section, empirically, for any given $L$ and $\beta$, we need to calculate the optima for the following constrained optimization problem, 
\begin{equation}
\label{pen_eq1}
\max_{R}\{\hat{E}[Y|\mathbf{X}\in R]\}
\end{equation}
subject to 
\begin{equation}
\label{pen_eq2}
\hat{P}_R > \beta,
\end{equation}
where $R=R_1 \bigcup...R_l...\bigcup R_L$, with $R_l=[r_{l1}^{l}, r_{l1}^{u}]\times...\times[r_{lj}^{l}, r_{lj}^{u}]\times...\times[r_{lp}^{l}, r_{lp}^{u}]$, $r_{lj}^{l}, r_{lj}^{u}\in \mathbb{R}$ and $0<r_{lj}^{l}< r_{lj}^{u}<\infty$. $\forall l, l^\prime \in \{1,...,L\}$ and $l \neq l^\prime$, there exist $j\in \{1,...,p\}$ such that \begin{equation}
\label{pen_eq3}
\max({r_{lj}^{l}, r_{l^\prime j^*}^{l}}) > \min({r_{lj}^{u}, r_{l^\prime j}^{u}}).
\end{equation}
Note that $\hat{E}[Y|\mathbf{X}\in R]$ and $\hat{P}_R$ are given by  Eq.~\eqref{est_eq1}, \eqref{est_eq2}.


Note that the above constrained optimization problem involves $2Lp$ decision variables (i.e., $r_{lj}^{l}, r_{lj}^{u}, l \in \{1,...,L\}$ and $j\in \{1,...,p\}$) and the disjoint constraint Eq.~\eqref{pen_eq3} actually involves $L \choose 2$$*p$ individual conditions. To further boost the computational efficiency, we define a new objective function with the disjoint constraint Eq.~\eqref{pen_eq3} considered as $O(R)$. Let the set consist of all regions that satisfy the requirements regarding region $R$ in Eq.~\eqref{pen_eq3} be $\Theta_{\mathbf{X}}$. The new objective function $O(R)$ is defined as
\begin{equation}\label{pen_eq11}
    O(R) = \left \{
  \begin{aligned}
    &\hat{E}[Y|\mathbf{X}\in R], && R \in \Theta_{\mathbf{X}} \\
    &\eta, && R \notin \Theta_{\mathbf{X}} 
  \end{aligned} \right.
\end{equation}
with $\eta$ being a numerical number that is smaller than the minimum value of $y_i$, $i=1,...,n$. The problem in Eq.~\eqref{pen_eq1}, Eq.~\eqref{pen_eq2} and Eq.~\eqref{pen_eq3} is simplified to 
\begin{equation}\label{pen_eq22}
\max_{R}\{O(R)\}.
\end{equation}
subject to 
\begin{equation}
\label{pen_eq33}
\hat{P}_R > \beta.
\end{equation}


Lagrangian relaxation is a popular approach to solve the constrained optimization problem like our problem expressed in Eq.~\eqref{pen_eq22} and Eq.~\eqref{pen_eq33}. However, the Lagrangian relaxation approach is known to suffer the problem of the duality gap under the general setting. It is not guaranteed that we always obtain the global or even local optima. In our numerical studies, we experimented the Lagrangian relaxation approach but failed to find the optimal solutions in some situations. In this paper, we propose to use an augmented Lagrangian method, called the penalty method. To be more specific, the original constrained optimization problem in Eq.~\eqref{pen_eq22} and Eq.~\eqref{pen_eq33} is equivalently transformed into 
\begin{equation}
\label{pen_eq4}
\max_{R}\{O(R) - c^*(\beta-\hat{P}_R)^+\}, 
\end{equation}
where $(v)^+=\max(v, 0)$, $c^*$ is the minimal point of function $K(c)$, $c>0$, defined as 
\begin{equation}
\label{pen_eq5}
K(c)=\max_{R}\{O(R)  - c(\beta-\hat{P}_R)^+\}.
\end{equation}

\subsection{Genetic algorithm for optimization}\label{sec3.4}
The next step is to build a module to solve the single objective optimization problem in Eq.~\eqref{pen_eq5} for any given $c > 0$. Due to the inaccessibility to the closed form of the objective function and the high dimensional nature (total number of parameters for any given $R$ is $2pL$), we propose to use the genetic algorithm (GA), one of the evolutionary algorithms, to conduct the optimization task \cite{deb2002fast}. GA does not require information or approximation on derivatives, which is not available in our problem setting. Moreover, the GA is more robust than coordinate cycling directed search methods as it maintains a population of potential solutions in each iteration of searching and bears less risk of being trapped in a local minimum.

GA tries to mimic the natural process and consists of three components: \textit{selection},  based on the fitness score, determining which individual candidates are chosen for the mating process and how many offspring each candidate produces; \textit{crossover}, an exchange is performed between some variables of the parents, producing two new individuals; and \textit{mutation},  for a few selected offspring, one variable is altered by a small perturbation\cite{deb2002fast}. When using GA to optimize over continuous variables, the real coding is used and the fitness of individual candidates is evaluated by the magnitude of the objective function\cite{chelouah2000continuous}. 

To solve the optimization problem in Eq.~\eqref{pen_eq5} using GA, for any given value of $c, L, \beta$, we proceed as follows:
\begin{itemize}
    \item Step 1: Initiate a candidate population of size $W$: for $w=1,...,W$, denote the $w$-th initialization as a $2Lp$-dimensional vector $[r_{l,j,w}^{l}, r_{l,j,w}^{u}]^T$, for $l=1,...,L; j=1,...,p$. The region spanned by these $2Lp$ values is denoted as $R_w$.
    \item  Step 2: Evaluate the fitness (i.e., how good it is in terms of maximizing the objective function in Eq.~\eqref{pen_eq5} ) of each initialization by evaluating the objective function $O(R_w)  - c(\beta-\hat{P}_{R_w})^+$. 
    \item  Step 3: Check whether certain stopping criteria is met. For instance, the criteria can be the increment in the fitness, compared to the previous population, is smaller than some threshold. If yes, output the best individual as the identified region $R^*$. Otherwise, move to the next step to produce a new population. 
    \item  Step 4: There are several steps in the new population producing process: 
    \begin{itemize}
        \item Select which individuals among the current population will produce offspring based on their fitness, i.e., the fitted values of the objective function $O(R)-c(\beta-\hat{P}_{R})^+$.Individuals with higher fitness values have higher chances to be selected. 
        \item Some of the $2Lp$ variables are exchanged between the selected parents to produce offspring. 
        \item The produced offspring are mutated by certain perturbation with a certain probability. 
        \item The produced new population is fed into step 2. 
    \end{itemize}
\end{itemize}
In this paper, we used the R package `GA' \cite{scrucca2013ga} to numerically implement the above procedures. The built-in parallel running option is adopted to increase the efficiency.

\subsection{Generalization}\label{sec3.2}

In supervised machine learning problems, generalization refers to the ability of an algorithm being effective for unseen data \cite{michie1994machine}. The generalization of machine learning algorithms is important as it ensures that a similar accuracy level can be achieved when the model is deployed to make predictions for new data sets. 


Analogously, in our operating envelope problem, it is also important to ensure that the achieved mean value of the response variable (i.e., the KPI) when applying the identified region $R^*$ to unseen data does not deviate much from the mean value of the response variable achieved in the training phase due to sampling error. To ensure the generalization of the identified operating envelope, we propose the following regularized objective function
\begin{equation}
\label{obj_eq1}
\hat{E}[Y|\mathbf{X}\in R] - \gamma \text{SD}(\hat{E}[Y|\mathbf{X}\in R]),
\end{equation}
where $\text{SD}(\hat{E}[Y|\mathbf{X}\in R])$ is the sample standard deviation of the achieved estimate $\hat{E}[Y|\mathbf{X}\in R]$, $\gamma > 0$ represents the size of regularization on the sample standard deviation part. The intuition is that if the sample standard error (i.e., variation across different data sets) of the achieved optimal average KPI within the detected hyper-rectangle is small, we have better chance to maintain the same optimal KPI level for any new data set. 

There is usually no closed form for $\text{SD}(\hat{E}[Y|\mathbf{X}\in R])$ in Eq.~\eqref{obj_eq1}. We propose to use the bootstrapping re-sampling technique to non-parametrically estimate this quantity \cite{mooney1993bootstrapping}. The detailed estimation procedure is summarized as follows. Let the total number of bootstrapping be $M$, 
\begin{itemize}
    \item  Step 1: For iteration $m = 1$ to $M$:
    \begin{itemize}
        \item Sample with replacement from the raw index \{1,...,n\}, and get another index set of length $n$, $\{d_{m,1},...,d_{m,i},...,d_{m,n}\}$.
        \item The new data set of size $n$ is $(y_{d_{m,i}}, \mathbf{x}_{d_{m,i}})$ for $i=1,...,n$.
        \item Compute
        \begin{equation}
            \hat{E}[Y|\mathbf{X}\in R]_m = \frac{\sum_{i=1}^n y_{d_{m,i}} I(\mathbf{x}_{d_{m,i}}\in R)}{\sum_{i=1}^n I(\mathbf{x}_{d_{m,i}}\in R)}\defeq \hat{E}_m.
        \end{equation}
    \end{itemize}
    \item Step 2: Compute and output 
    \begin{equation}
        \hat{\text{SD}}_{\text{bs}}(\hat{E}[Y|\mathbf{X}\in R])=\sqrt{\frac{1}{M-1}\sum_{m=1}^M(\hat{E}_m-\bar{\hat{E}}_m)^2}.
    \end{equation}
\end{itemize}

Let the estimated sample standard deviation be $\hat{\text{SD}}_{\text{bs}}(\hat{E}[Y|\mathbf{X}\in R])$. The empirical regularized objective function for any given region $R$ is then 
\begin{equation}
\label{obj_eq2}
\hat{E}[Y|\mathbf{X}\in R] - \gamma \hat{\text{SD}}_{\text{bs}}(\hat{E}[Y|\mathbf{X}\in R]).
\end{equation}

 
The counterparts of the bias and variance concepts in supervised machine learning problems exist in the proposed operating envelope problem in Definition 1. Specifically, we would like to achieve as high KPI value as possible in our setting. Hence, we define the \textit{bias} as the gap between the expectation of the achieved KPI values for an unseen data set and the \textit{true optimal KPI value}. The \textit{true optimal KPI value} refers to the optimal objective value in Definition 1 assuming that we could have had access to the whole data population. In informal notation, $\text{Bias} = \text{True Optimal KPI}- E[\hat{E}^*_{\text{test}}[Y|\mathbf{X}\in R]]$. Since we would not have the \textit{true optimal KPI value}, we propose to use another non-increasing function of the expected KPI value of unseen data to represent bias. In particular,    

\begin{equation}
\label{bias}
   \text{Bias}=\frac{1}{E[\hat{E}^*_{\text{test}}[Y|\mathbf{X}\in R]]}.
\end{equation}

The \textit{variance} concept here is similar to its counterpart in machine learning. In particular, we define \textit{variance} as the variability of achieved KPI values of an unseen data set when applying the identified regions by solving the problem in Definition 1 using different training data sets. It is formally defined as follows:   
 
\begin{equation}
\label{var}
\text{Variance}=\text{Var}{[\hat{E}^*_{\text{test}}[Y|\mathbf{X}\in R]]}.
\end{equation}

For any given regularization parameter $\gamma$, the bias and variance defined above can be estimated by the cross-validation technique \cite{kohavi1995study}. Similar to supervised machine learning, there exists a trade-off between the bias and variance, which requires judgments of the real needs to decide how to balance between bias and variance in the operating envelope problem. This fact is demonstrated by the simulation studies and the real-world data analysis in Section \ref{sec4}. 


All the components discussed in Section \ref{sec3.1} to \ref{sec3.2} works together and forms the proposed regularized `GA + penalty' algorithm summarized in Algorithm 1.

\begin{algorithm}[htb]
  \caption{Regularized `GA + penalty' algorithm}
  \begin{enumerate}
    \item
   Achieve the desired regularization size $\gamma^*$:
    \begin{enumerate}
      \item
      Split data into training and testing sets.
      \item
      Specify a set of candidates for the regularization parameter $\gamma$, denoted as $\{\gamma_{1},....,\gamma_{k},...,\gamma_{K}\}$. 
     
      \item
      For any given $\gamma_{k}$, 
      \begin{enumerate}
           \item Specify a set of candidates for the penalty parameter $c$ in Eq.~\eqref{pen_eq5}, denoted as $\{c_{k,1},....,c_{k,t},...,c_{k,T_k}\}$. 
           \item For any given $c_{k,t}$, 
           \begin{itemize}
              \item Use training data to calculate $\hat{E}[Y|\mathbf{X}\in R]$.
              \item Use training data to bootstrap to achieve $\hat{\text{SD}}_{\text{bs}}(\hat{E}[Y|\mathbf{X}\in R])$.
              \item Use the genetic algorithm to solve 
              \begin{multline}
              \max_R\{\hat{E}[Y|\mathbf{X}\in R] - \gamma_{k}\hat{\text{SD}}_{\text{bs}}(\hat{E}[Y|\mathbf{X}\in R])\\ 
              - c_{k,t}(\beta-\hat{P}_R)^+\}\defeq H_{train,k,t}. 
              \end{multline}
              \end{itemize}
            \item The desired penalty parameter for $\gamma_{k}$ is set as 
            \begin{equation}
                c_{k}^* = \min_{t}H_{train,k,t}
            \end{equation}
             The corresponding optimal region $R_k^*$ is saved. 
             \item Use $R_k^*$ to calculate the bias and variance term corresponding to $\gamma_{k}$ with testing data.
      \end{enumerate}
      \item
      Based on the bias and variance over different $\gamma_{k}$ to determine the desired $\gamma^*$. 
      
    \end{enumerate}

  \item
  Output the region detected corresponding to $\gamma^*$ stored in step c) as the optimal operating envelope.
  \end{enumerate}
\end{algorithm}

\section{Numerical Experiments}\label{sec4}

\subsection{Simulation study I} \label{sec4.1}
In this subsection, we consider a set of simulations with $p=1$ state variable to demonstrate the effectiveness of our proposed regularized `GA + penalty' algorithm in Section \ref{sec3} for the operating envelope with interpretability and implementability constraints problem defined in Section \ref{sec2}. The first two experiments focus on investigating the impacts of different components in Definition 1 on the region detection results. In these two experiments, the variability is simulated to be the same across different sub-regions of $\Omega(\mathbf{X})$ and the `GA + penalty' algorithm without regularization is utilized. In the latter two experiments, we investigate the benefit of adding regularization. For all the simulation studies, we assume that data $(y_i, x_i)$, for $i=1,...,n$ are generated from model:
\begin{equation}
\label{sim_eq1}
y_i = f(x_i) + \epsilon_i, 
\end{equation}
where $x_i \sim \text{i.i.d } g(X)$ and $\epsilon_i \sim \text{i.i.d } N(0, \sigma_{\epsilon}^2)$. 

\subsubsection{Simulation I-(a)}
\begin{figure}[htbp]
	\centering
	\begin{subfigure}[t]{1.35in}
		\centering
        \caption{$\delta=0$}
        \vspace{-0.1in}
		\includegraphics[width=32mm]{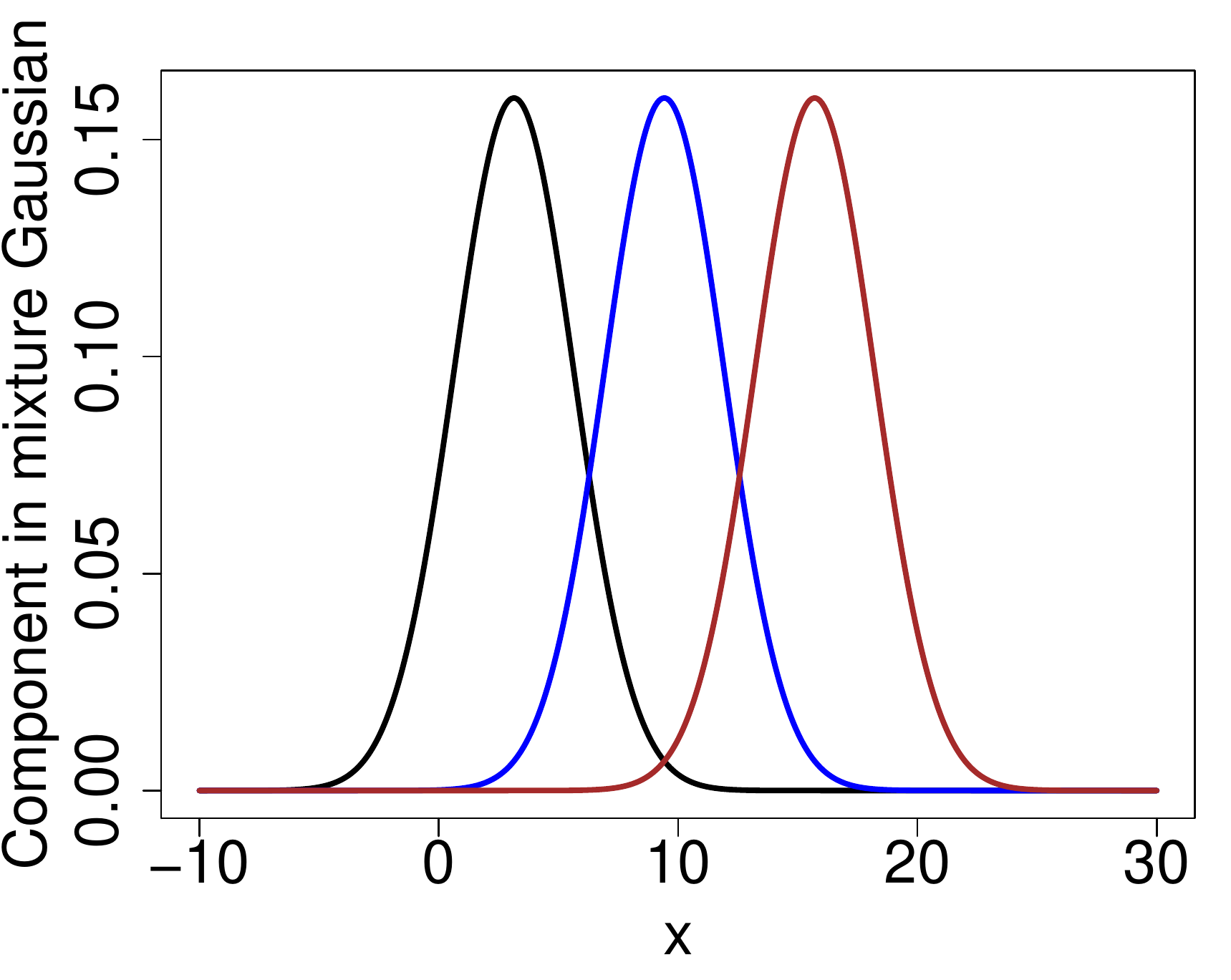} \label{g1}	
	\end{subfigure}
	\quad
	\begin{subfigure}[t]{1.35in}
		\centering
        \caption{$\delta=0.3$}
        \vspace{-0.1in}
		\includegraphics[width=32mm]{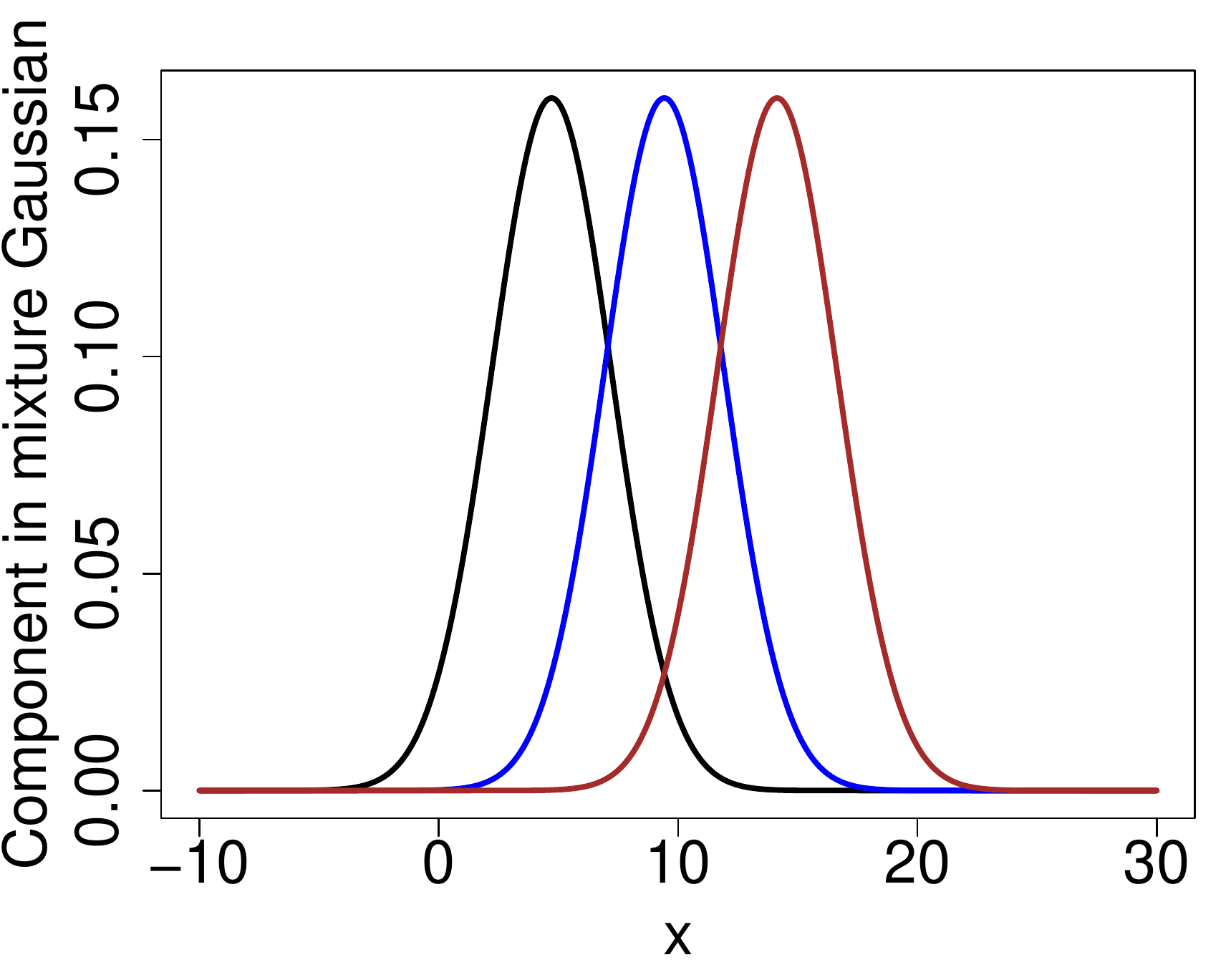} 		
	\end{subfigure}
    \par \bigskip
    \vspace{-0.1in}
	\begin{subfigure}[t]{1.35in}
		\centering
        \vspace{-0.1in}
		\includegraphics[width=32mm]{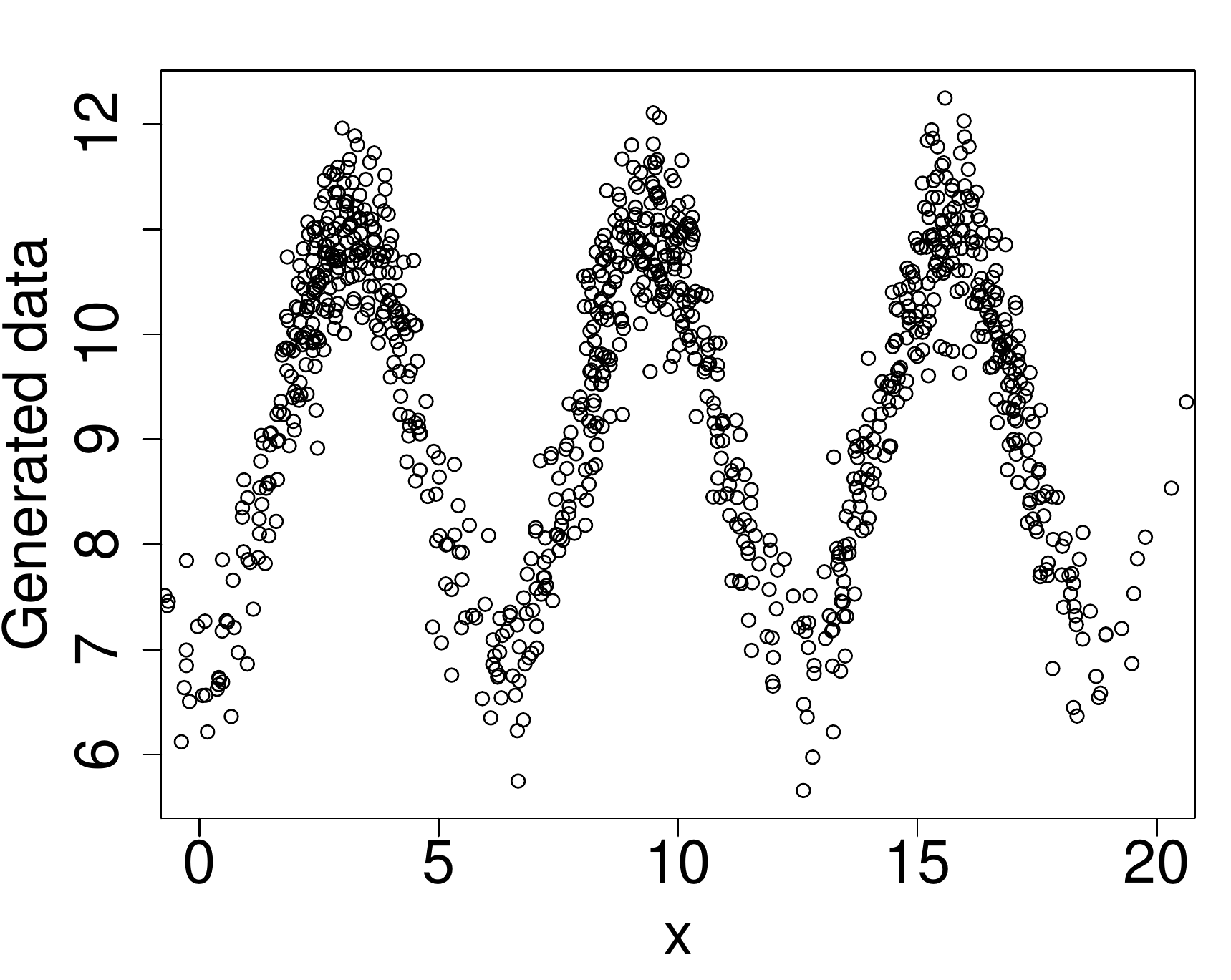} 		
	\end{subfigure}
	\quad
	\begin{subfigure}[t]{1.35in}
		\centering
        \vspace{-0.1in}
		\includegraphics[width=32mm]{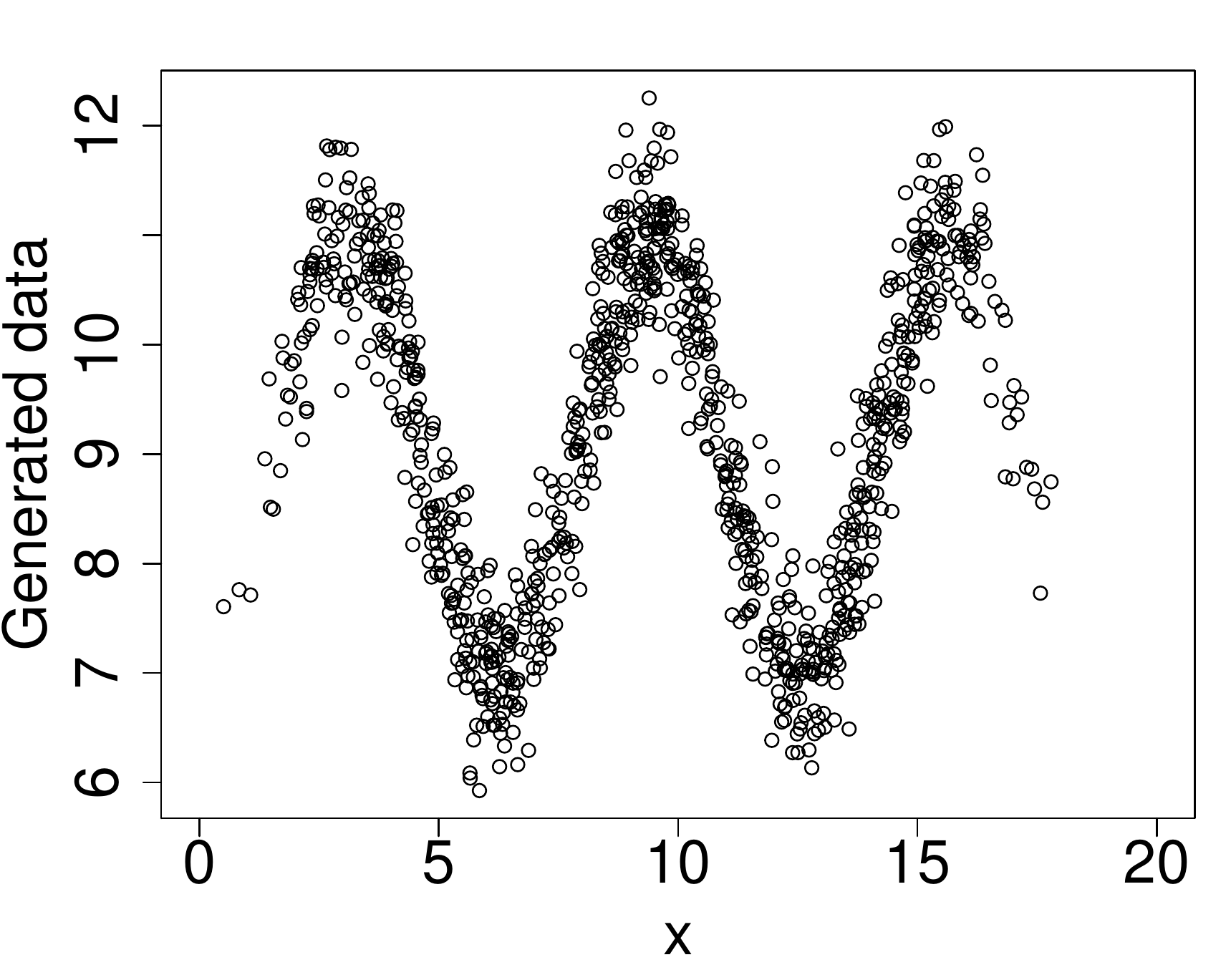} 		
	\end{subfigure}
	\caption{Components of the mixture Gaussian density $g(X)$ and the generated data with $\delta=0, 0.3$ in Simulation I-(a). }\label{sim1a-1}
\end{figure}

In this simulation, we demonstrate the influence of $g(X)$ on the operating envelope output. The mapping function is $f(X)= -2(\cos(X)-4.5)$ (see Fig.~\ref{set1}). The number of samples is $n=1000$. The standard deviation of errors is $\sigma_{\epsilon}=0.25$. The probability density function $g(X)$ is a mixture of three Gaussian distributions, i.e.,  it with probability $1/3$ of being $N(\mu_s, 2.5)$, for $s=1,2,3$. The mean values are $\mu_1=(1+\delta)\pi$, $\mu_2=3\pi$, and $\mu_3=(5-\delta)\pi$, with parameter $0\leq\delta \leq 1$ quantifying the sparseness of data around the first and the third peaks of $f(X)$. The Gaussian distributions and the corresponding generated data for $\delta=0$ and $\delta=0.3$ are provided in Fig.~\ref{sim1a-1}. As shown by Fig.~\ref{sim1a-1}, with the increase of $\delta$, more data are centered closed to the middle peak $\mu_2=3\pi$. 

\begin{figure}[htbp]
	\centering
	\begin{subfigure}[t]{0.95in}
		\centering
        \caption{$\delta=0$}
        \vspace{-0.1in}
		\includegraphics[width=25mm, height=23.5mm]{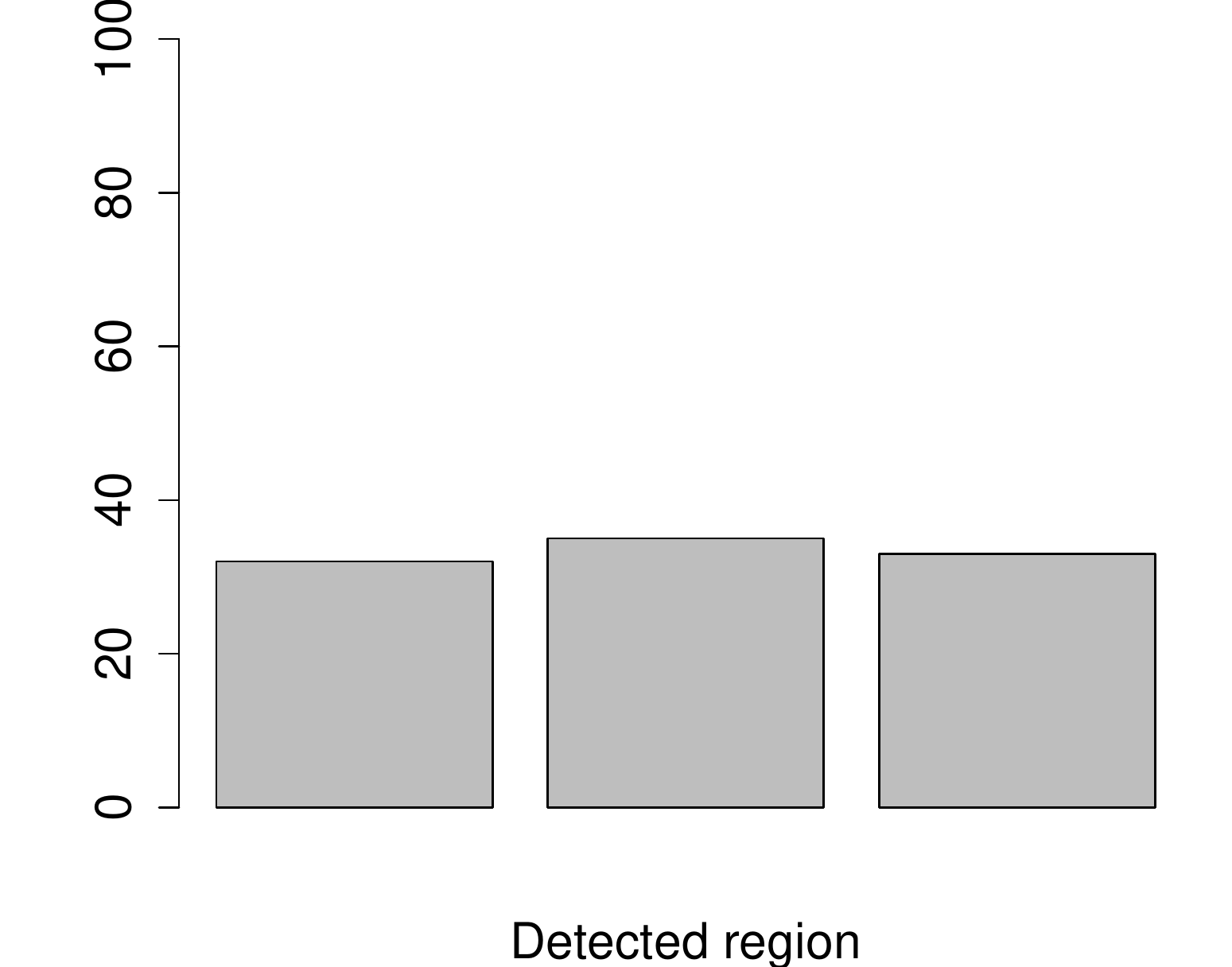} \label{bar1}	
	\end{subfigure}
	\quad
	\begin{subfigure}[t]{0.95in}
		\centering
        \caption{$\delta=0.1$}
        \vspace{-0.1in}
		\includegraphics[width=25mm, height=23.5mm]{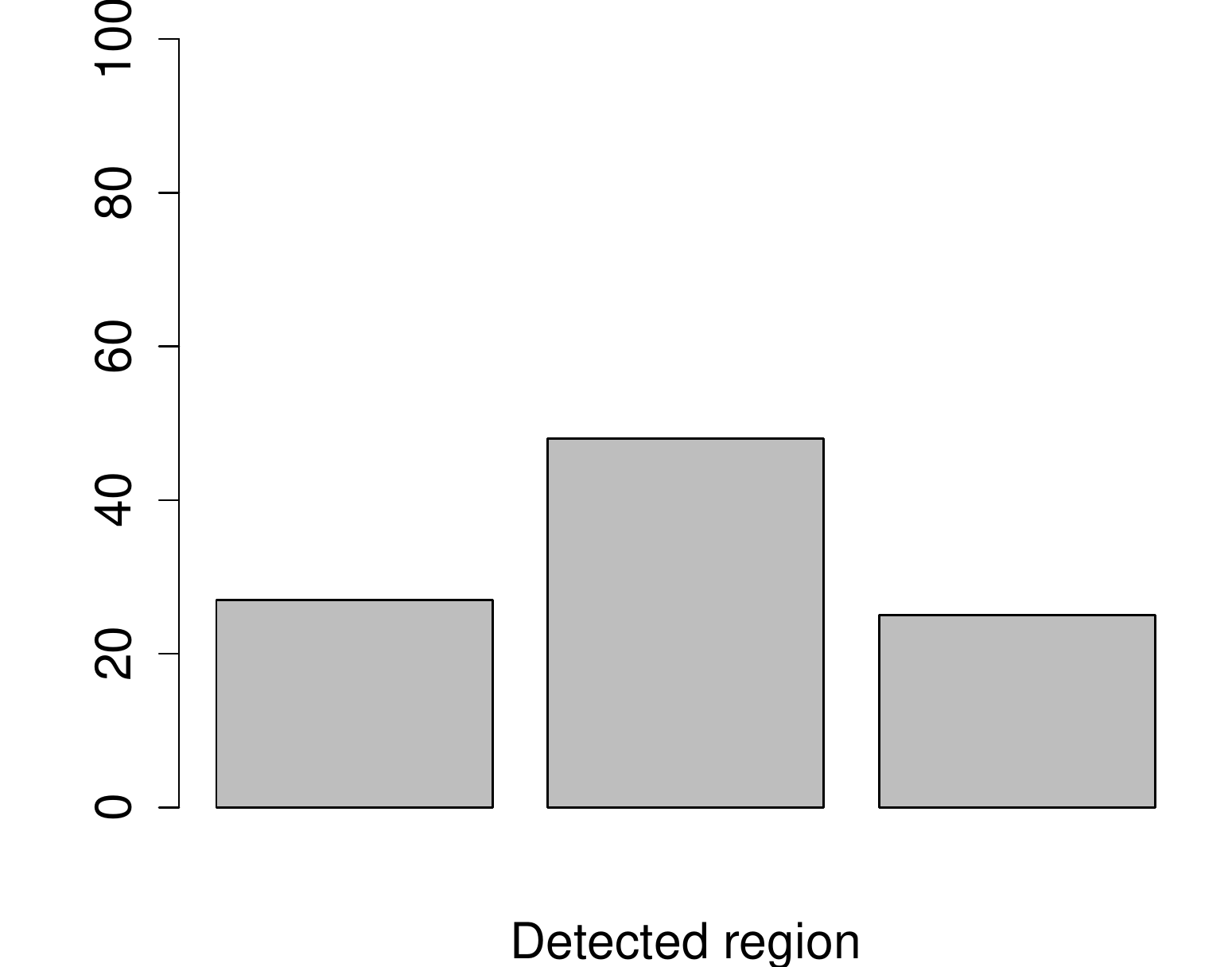} 		
	\end{subfigure}
	\quad
	\begin{subfigure}[t]{0.95in}
		\centering
        \caption{$\delta=0.3$}
        \vspace{-0.1in}
		\includegraphics[width=25mm, height=23.5mm]{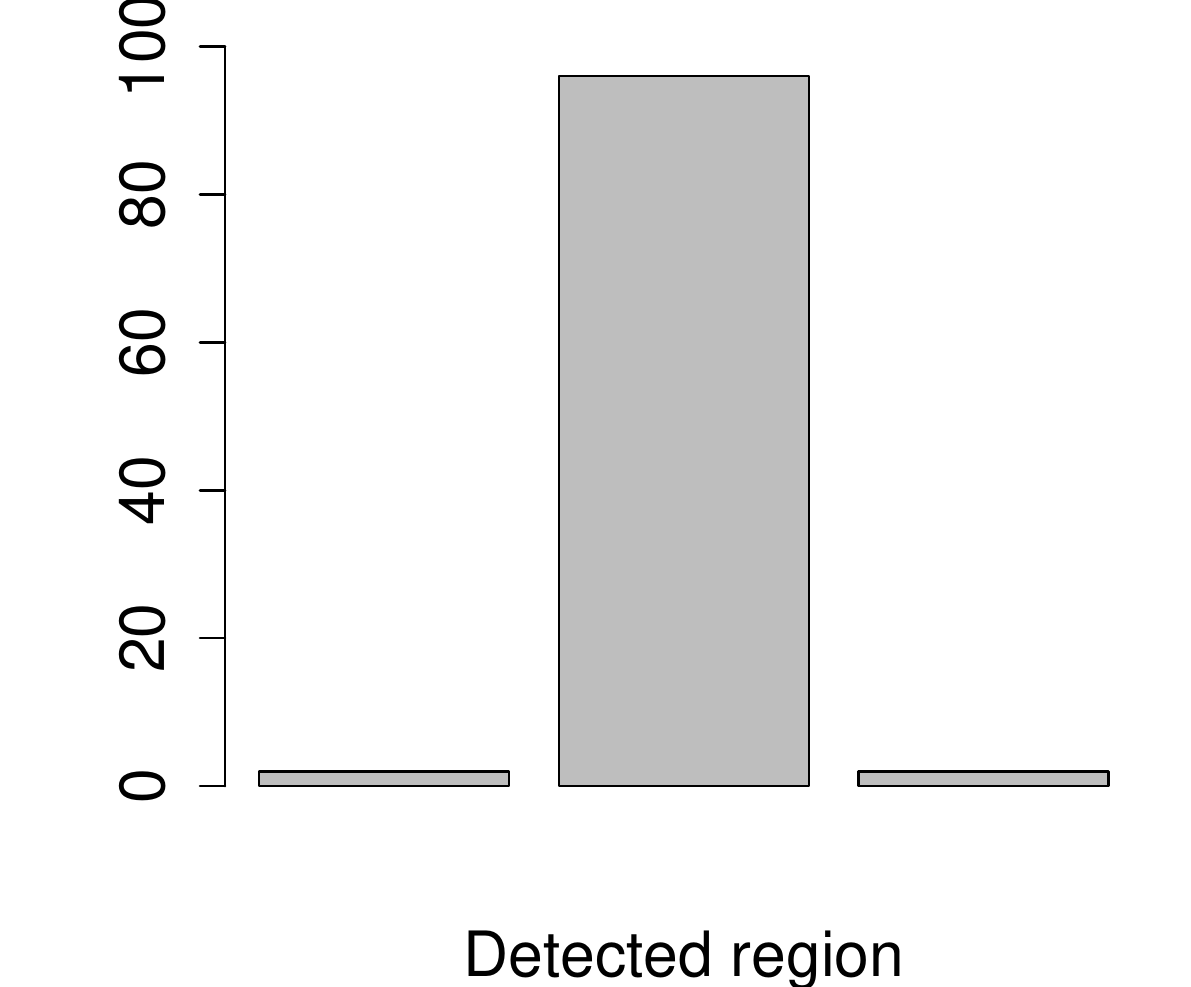} 		
	\end{subfigure}
	\vspace{-0.2in}
	\caption{Simulation I-(a). Bar charts for the location of the detected region under different probability density functions, i.e. $\delta=0, 0.1, 0.3$. The three bars represent the number of times that the detected region being within $[0, 2\pi)$, $[2\pi, 4\pi)$, and $[4\pi, 6\pi]$ respectively over 100 Monte Carlo repetitions.}\label{sim1a-2}
\end{figure}

The probabilistic coverage threshold is specified as $\beta=0.25$. The number of disjoint regions is $L=1$. Three different values for $\delta$ are used, i.e., $\delta=0, 0.1, 0.3$. For a given $\delta$, we use the proposed `GA + penalty' solution in Section \ref{sec3} to calculate the operating envelope (i.e., interval) that satisfies the probabilistic coverage constraint. The calculation is repeated for $100$ Monte Carlo simulations. The frequencies that the calculated interval falling into each of the three sub-regions $[0, 2\pi)$, $[2\pi, 4\pi)$, and $[4\pi, 6\pi]$ are visualized in the bar charts in Fig.~\ref{sim1a-2}. Our proposed algorithm produces reasonable results. With the increase of $\delta$, i.e., more data are centered around the middle region, the proportion of the identified operating envelopes falling into the middle sub-region increases.

\subsubsection{Simulation I-(b)}
In this experiment, we investigate the impact of the probabilistic coverage threshold $\beta$. The state variable $X$ is distributed the same as the one in the previous experiment with $\delta=0$. $f(X)$ is
\begin{equation}\label{sim1b}
    f(X) = \left \{
  \begin{aligned}
    &-1.95(\cos(X)-4.5),  && 0 \leq X < 2\pi \\
    &-2(\cos(X)-4.5), && 2\pi \leq X < 4\pi\\
    &-1.9(\cos(X)-4.5), && 4\pi \leq X \leq 6\pi
  \end{aligned} \right. .
\end{equation}
The standard deviation of random errors is $\sigma_{\epsilon}=0.25$. 

The calculated operating envelope for probabilistic coverage threshold $\beta=0.2, 0.4, 0.6$ are given in Fig.~\ref{sim1b}. The output from our proposed algorithm is well aligned with our intuition that the calculated operating envelope is wider when the required probabilistic coverage is larger.

\begin{figure}[htbp]
	\centering
	\begin{subfigure}[t]{0.98in}
		\centering
        \caption{$\beta=0.2$}
        \vspace{-0.1in}
		\includegraphics[width=27mm, height=24mm]{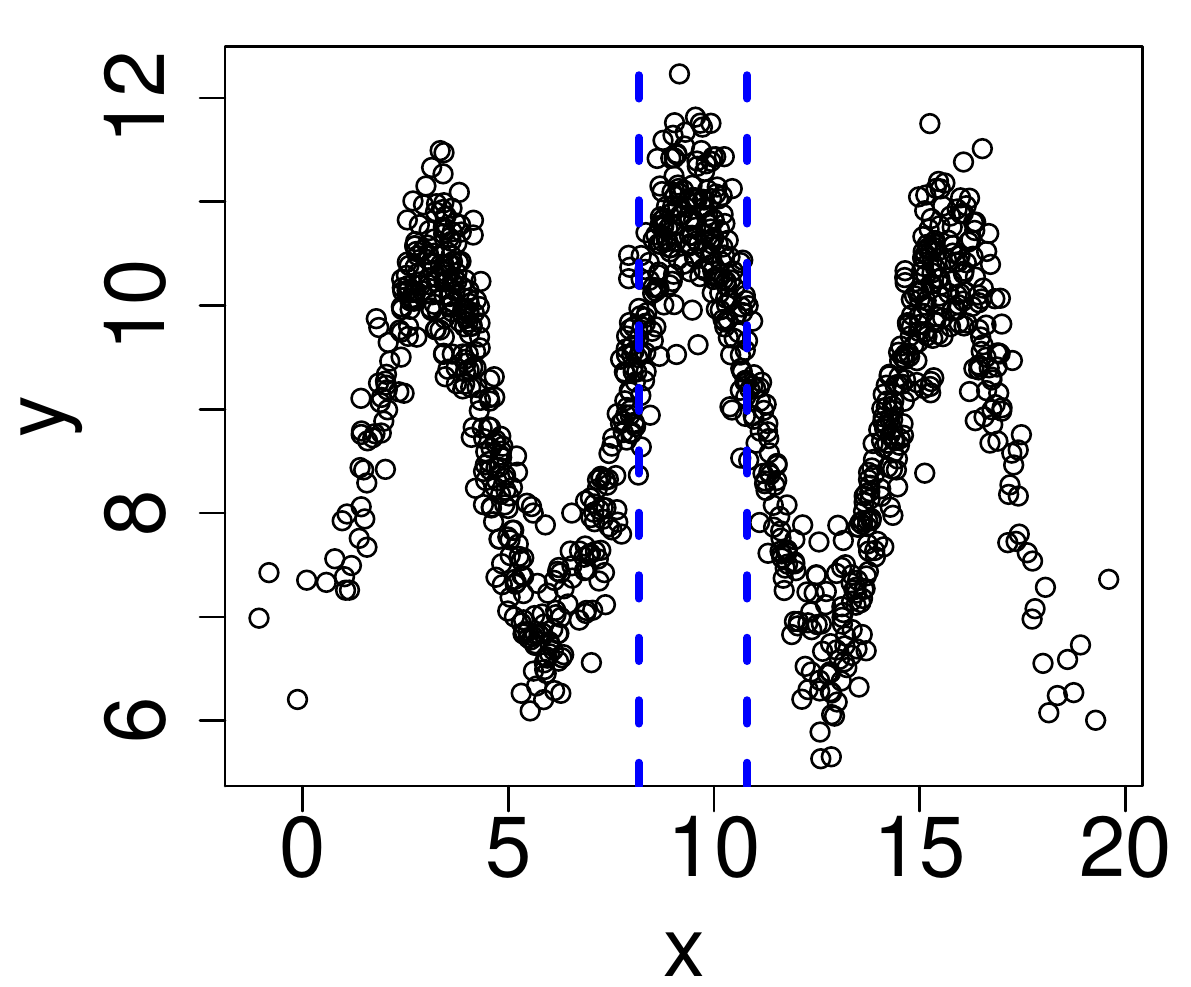} 		
	\end{subfigure}
	\quad
	\begin{subfigure}[t]{0.98in}
		\centering
        \caption{$\beta=0.4$}
        \vspace{-0.1in}
		\includegraphics[width=27mm, height=24mm]{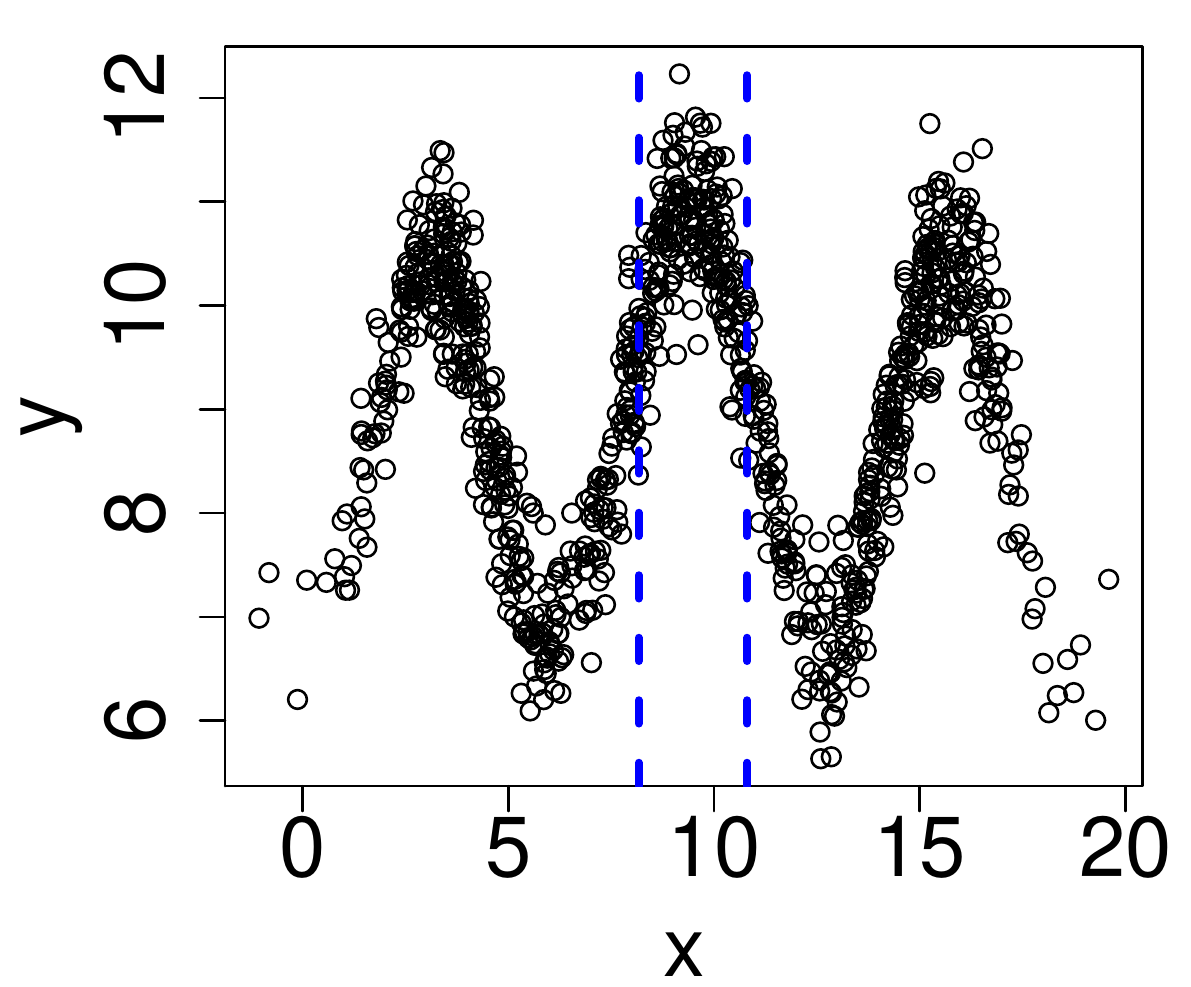} 		
	\end{subfigure}
	\quad
	\begin{subfigure}[t]{0.98in}
		\centering
        \caption{$\beta=0.6$}
        \vspace{-0.1in}
		\includegraphics[width=27mm, height=24mm]{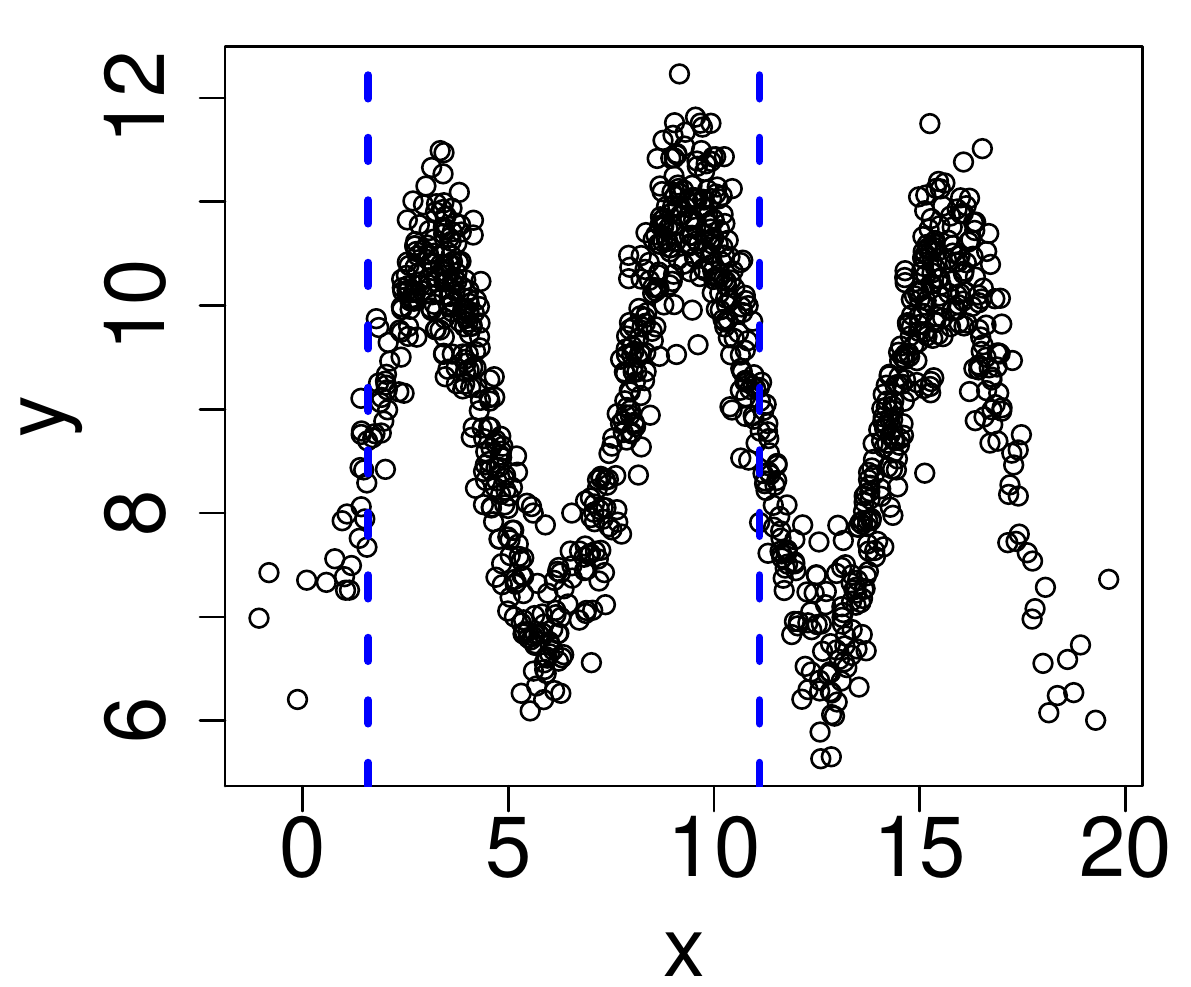} 		
	\end{subfigure}
	\caption{Simulation I-(b). Calculated operating envelope indicated by the two vertical lines for probabilistic coverage threshold $\beta=0.2, 0.4, 0.6$.}\label{sim1b}
\end{figure}

\subsubsection{Simulation I-(c)}
\begin{figure}[htbp]
	\centering
	\begin{subfigure}[t]{1.35in}
		\centering
        \caption{Assumed $f(X)$.}\label{set1}
        \vspace{-0.1in}
		\includegraphics[width=32mm]{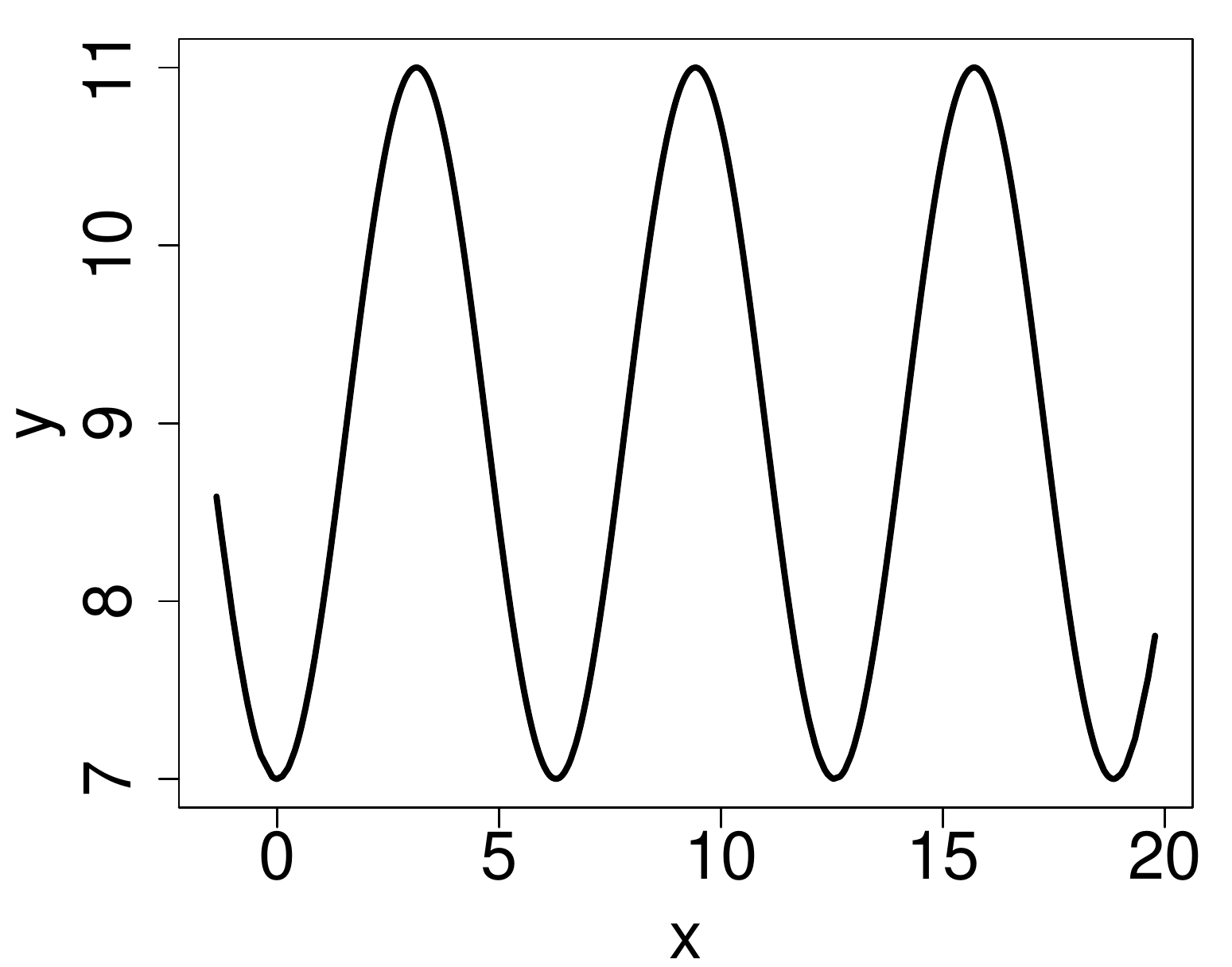} 		
	\end{subfigure}
	\quad
	\begin{subfigure}[t]{1.35in}
		\centering
        \caption{Components of $g(X)$.}\label{set2}
        \vspace{-0.1in}
		\includegraphics[width=32mm]{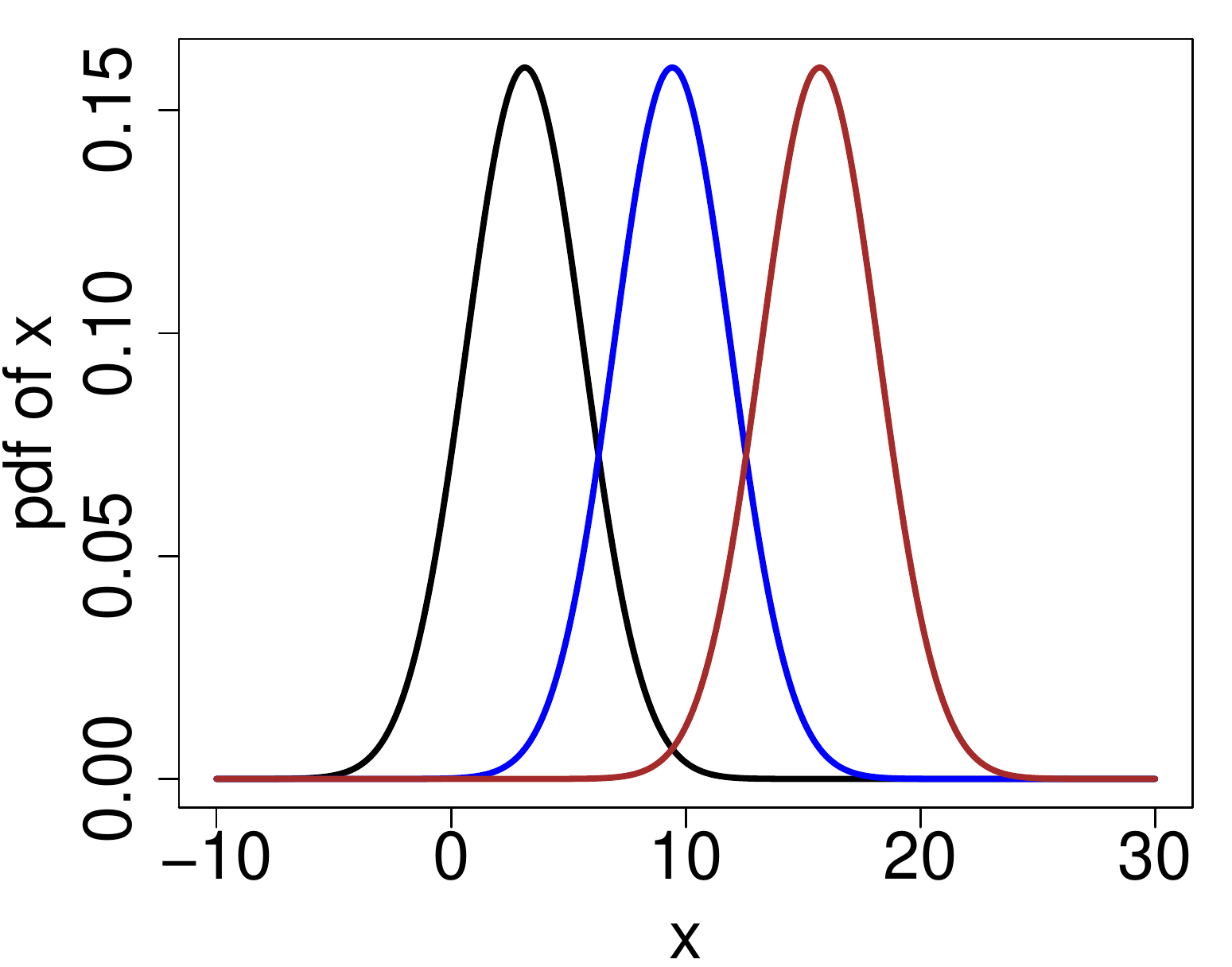} 		
	\end{subfigure}
    \par \bigskip
    \vspace{-0.1in}
	\begin{subfigure}[t]{1.35in}
		\centering
        \caption{Standard deviation of error at different $X$. }\label{set3}
        \vspace{-0.1in}
		\includegraphics[width=32mm]{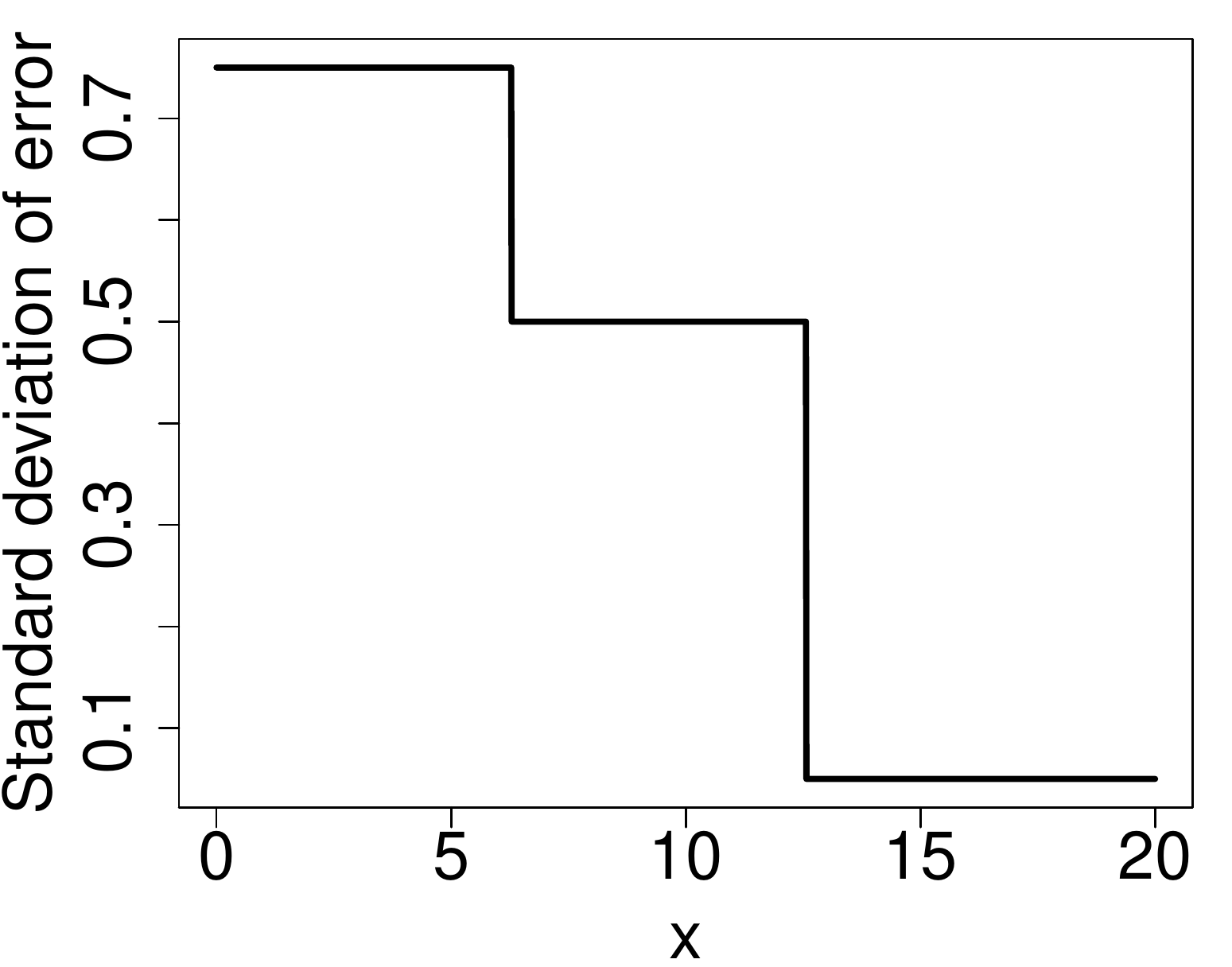} 		
	\end{subfigure}
	\quad
	\begin{subfigure}[t]{1.35in}
		\centering
        \caption{Generated $n=1000$ data samples, $(x_i, y_i)$.}\label{set4}
        \vspace{-0.1in}
		\includegraphics[width=32mm]{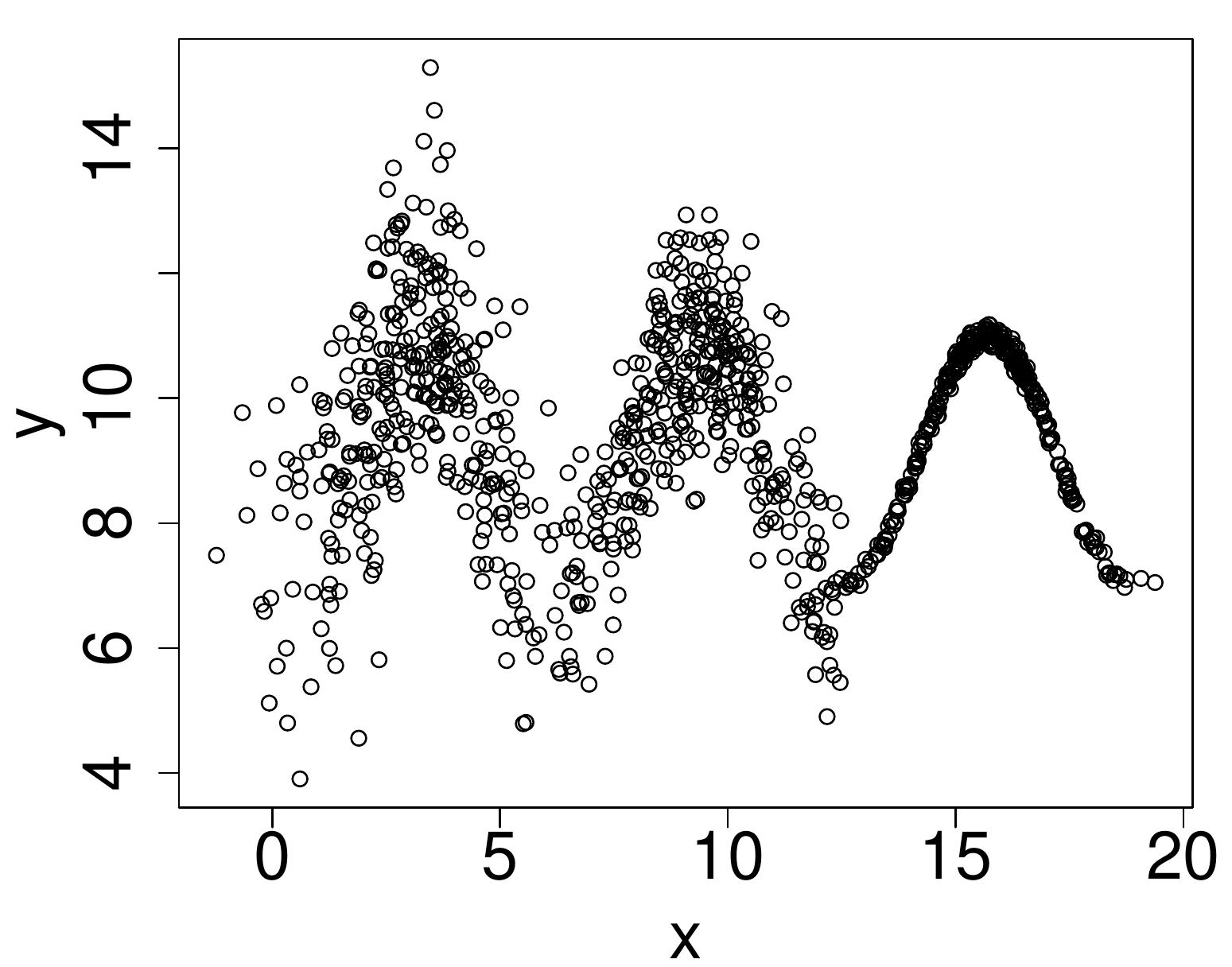} 		
	\end{subfigure}
	\caption{Settings and the generated data for Simulation I-(c). }\label{sim1c-1}
\end{figure}

In this study, we demonstrate the benefit of adding regularization.  The state variable $X$ is the same as the Simulation I-(b). The relation between $X$ and $Y$ is quantified by $f(X)=-2(\cos(X)-4.5)$, which is the same as the first simulation study (Fig.~\ref{set1}). The standard deviation of random errors takes three values, $\epsilon=0.75, 0.5, 0.05$, for the three sub-regions,  $[0, 2\pi)$, $[2\pi, 4\pi)$, and $[4\pi, 6\pi]$ respectively (Fig.~\ref{set3}). The generated data are plotted in Fig.~\ref{set4}. Based on the simulation setting, it can be seen that sub-region $[4\pi, 6\pi]$ is a better region to construct an operating envelope. This is because the magnitude of $\hat{E}[Y|\mathbf{X}\in R]$ are the same among all three sub-regions and the sampling standard deviation of $\hat{E}[Y|\mathbf{X}\in R]$ is smaller when $R\in [4\pi, 6\pi]$, which means that the evaluated value of $\hat{E}[Y|\mathbf{X}\in R]$ for any new data set has a larger chance to be close to the current value. The probabilistic coverage threshold considered is $\beta=0.25$. 

\begin{figure}[htbp]
	\centering
	\begin{subfigure}[t]{1.35in}
		\centering
        \caption{Without regularization}\label{woreg}
        \vspace{-0.1in}
		\includegraphics[width=33mm]{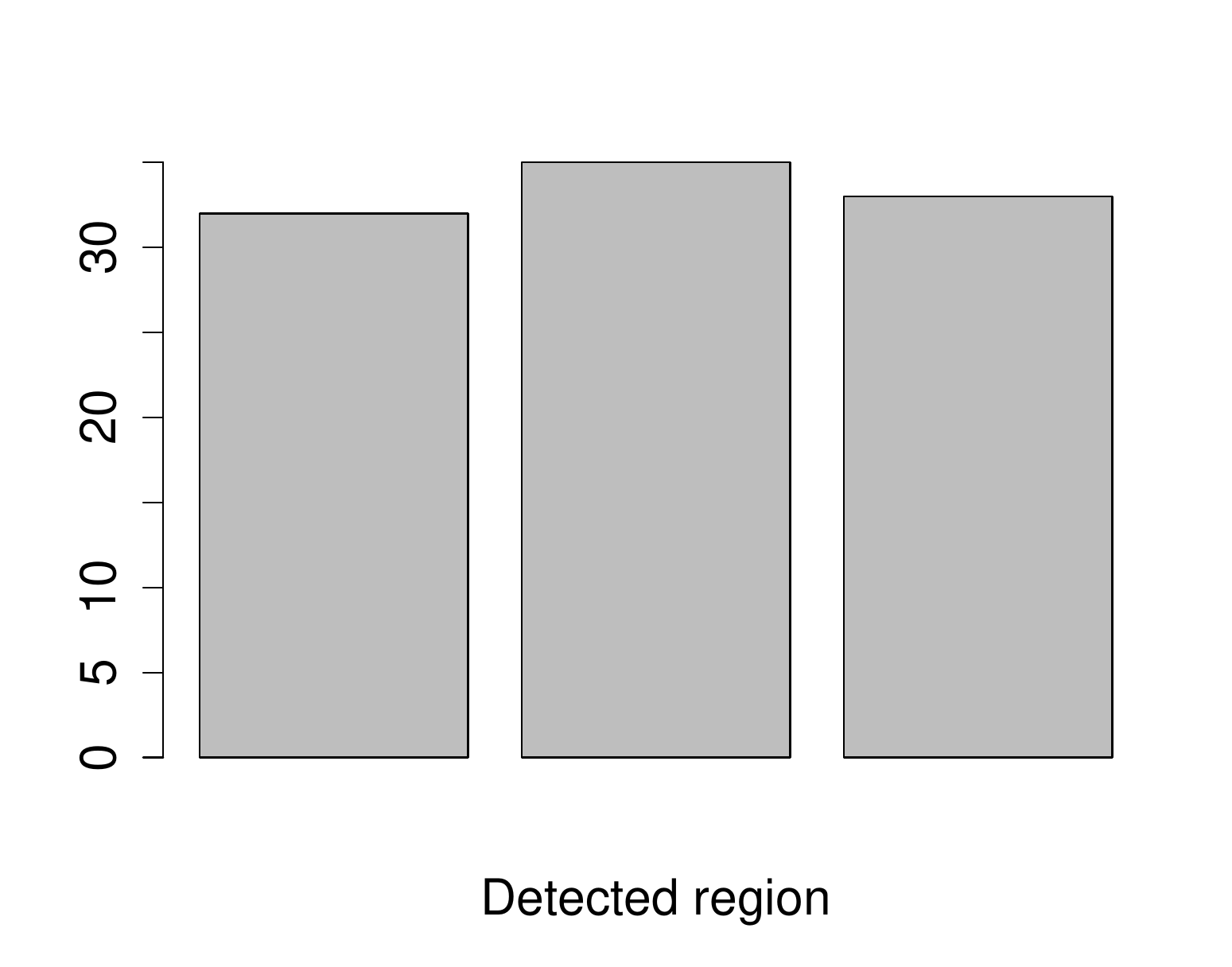} 		
	\end{subfigure}
	\quad
	\begin{subfigure}[t]{1.35in}
		\centering
        \caption{With regularization} \label{wreg}
        \vspace{-0.1in}
		\includegraphics[width=33mm]{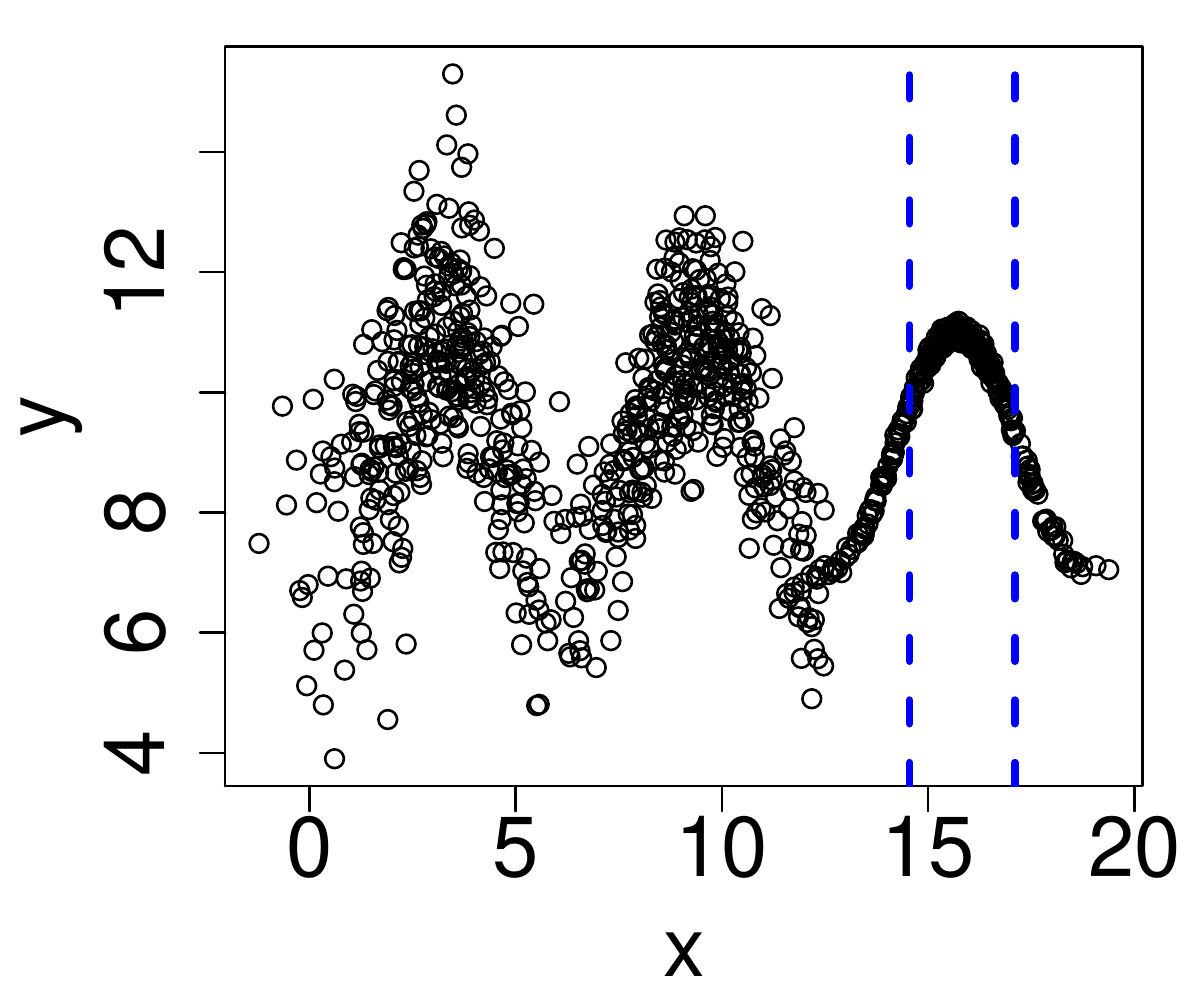} 		
	\end{subfigure}
	\caption{Simulation I-(c). Left: Bar chart for the location of the detected region without using regularization. Right: The achieved envelope with regularization ($\gamma=5.5$).}
\end{figure}

\begin{figure}[htbp]
	\centering
	\begin{subfigure}[t]{1.35in}
		\centering
        \caption{Simulation I-(c)}\label{sel1}
		\includegraphics[width=36mm, height=24.5mm]{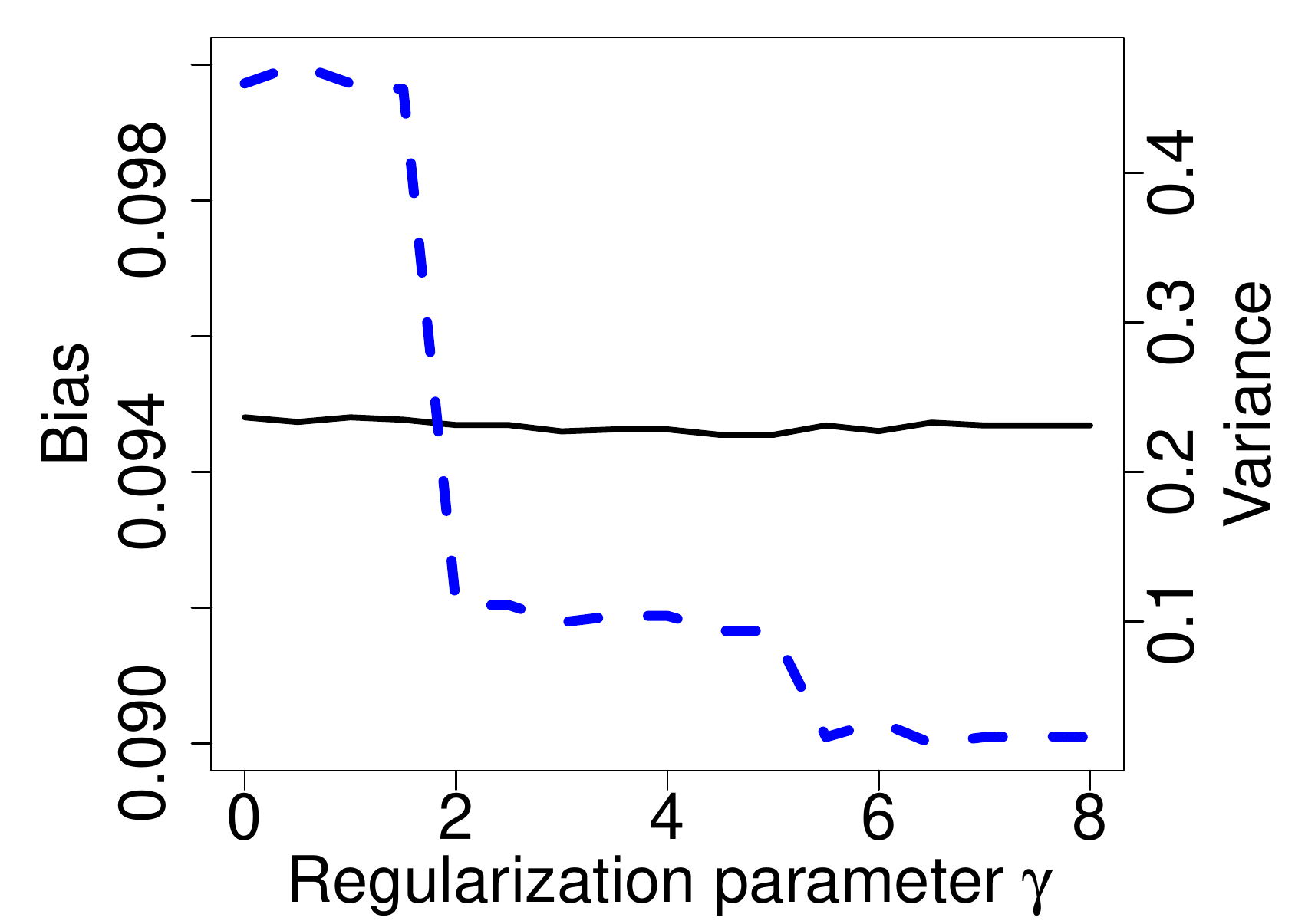} 		
	\end{subfigure}
	\quad
	\begin{subfigure}[t]{1.35in}
		\centering
        \caption{Simulation I-(d)}\label{sel2}
		\includegraphics[width=36mm, height=24.5mm]{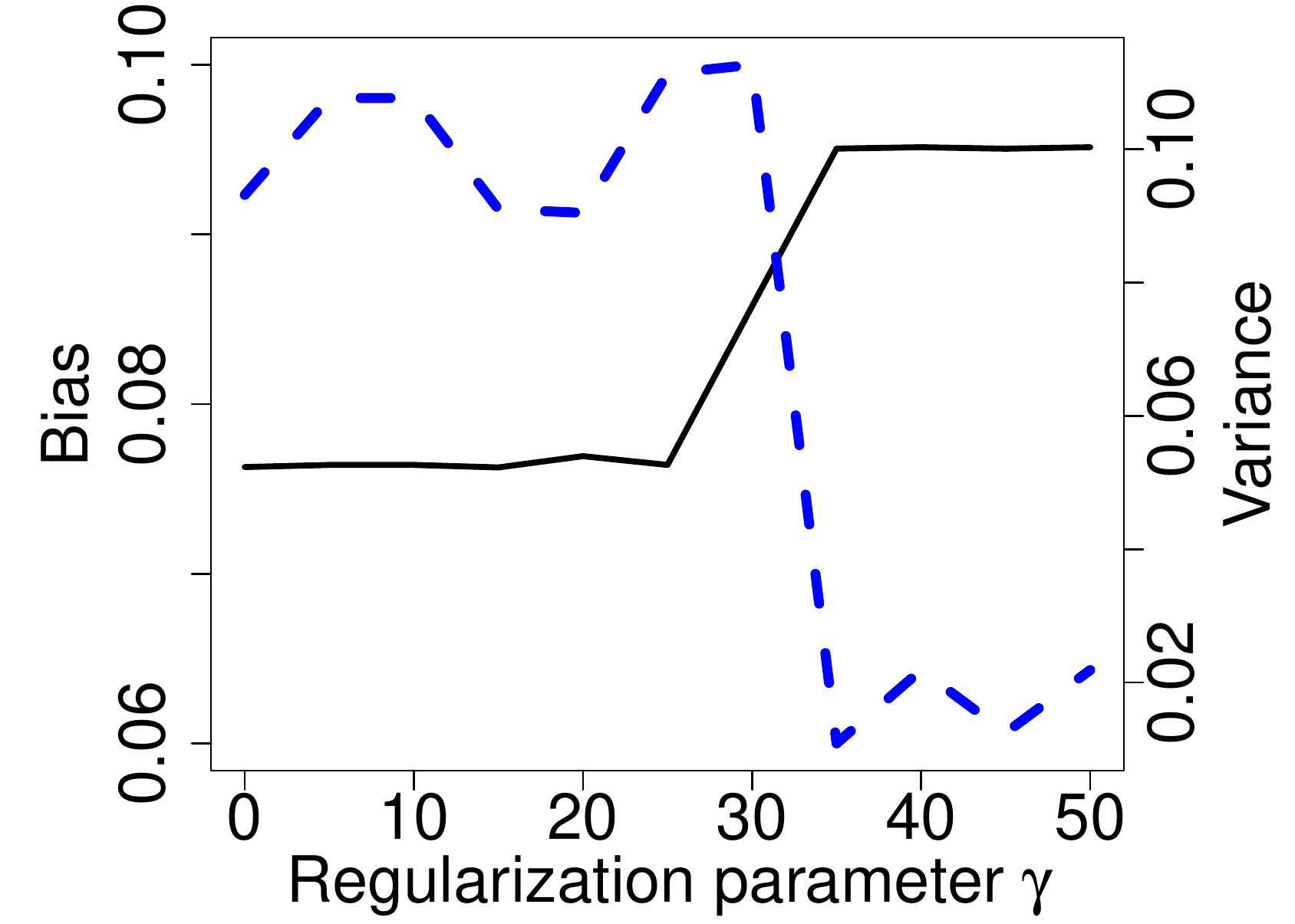} 		
	\end{subfigure}
	\caption{Bias (black line) and variance (blue line) as a function of the regularization parameter $\gamma$ in Simulation I-(c),(d).}\label{sim1c-2}
\end{figure}

The three sub-regions have the same $f(X)$ and $g(X)$. Without taking the variability of relevant statistics into account, the detected operating envelope has the same chance to be within any of these three sub-regions. This phenomenon is justified by Fig.~\ref{woreg}. Now we apply the regularized `GA + penalty' technique described in Section \ref{sec3.2}. For any given region, the sample standard deviation of $\hat{E}[Y|\mathbf{X}\in R]$ is done by the bootstrapping procedure with $M=500$. For any candidate $\gamma$ among $\{0, 0.5, 1, 1.5, ...,7.5, 8\}$, the 4-fold cross-validation is used to estimate the bias and variance in Eq.~\eqref{bias} and Eq.~\eqref{var}. According to Fig.~\ref{sel1}, with the increase of $\gamma$, the black bias curve (i.e., the evaluated mean response within the detected region) stays relatively stable, while the blue variance curve (.e., the variability of achieved optimal mean KPI) declines. We set $\gamma=5.5$, the smallest $\gamma$ that makes variance reach the minimum. The corresponding optimal region identified by the proposed algorithm is visualized in Fig.~\ref{wreg}. Our proposed regularization term effectively accounts for the variability factor.

\subsubsection{Simulation I-(d)}
\begin{figure}[htbp]
	\centering
	\begin{subfigure}[t]{1.45in}
		\centering
        \caption{Assumed $f(X)$.}\label{set11}
        \vspace{-0.1in}
		\includegraphics[width=33mm]{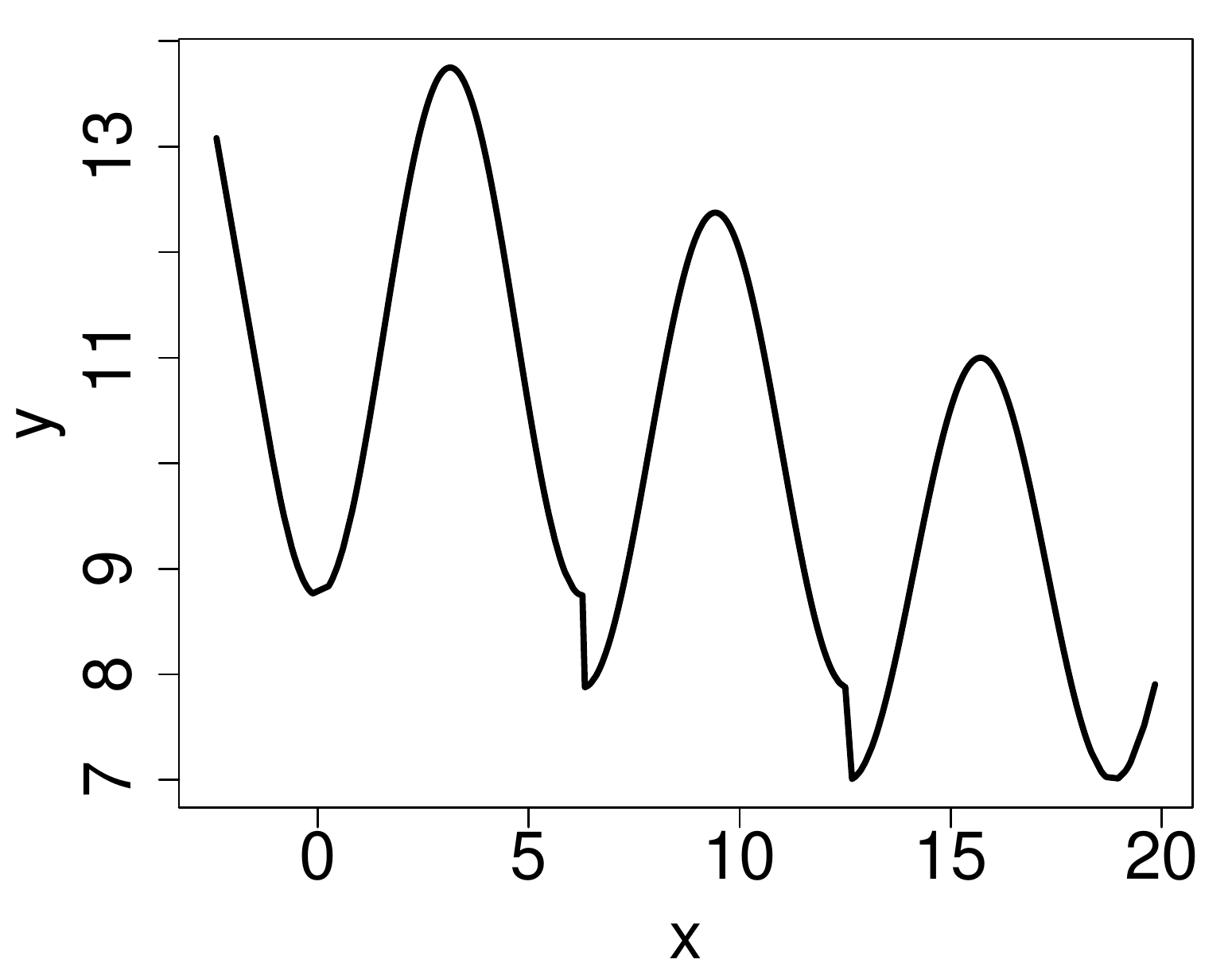} 		
	\end{subfigure}
	\quad
	\begin{subfigure}[t]{1.45in}
		\centering
        \caption{Generated data samples.}\label{set44}
        \vspace{-0.1in}
		\includegraphics[width=33mm]{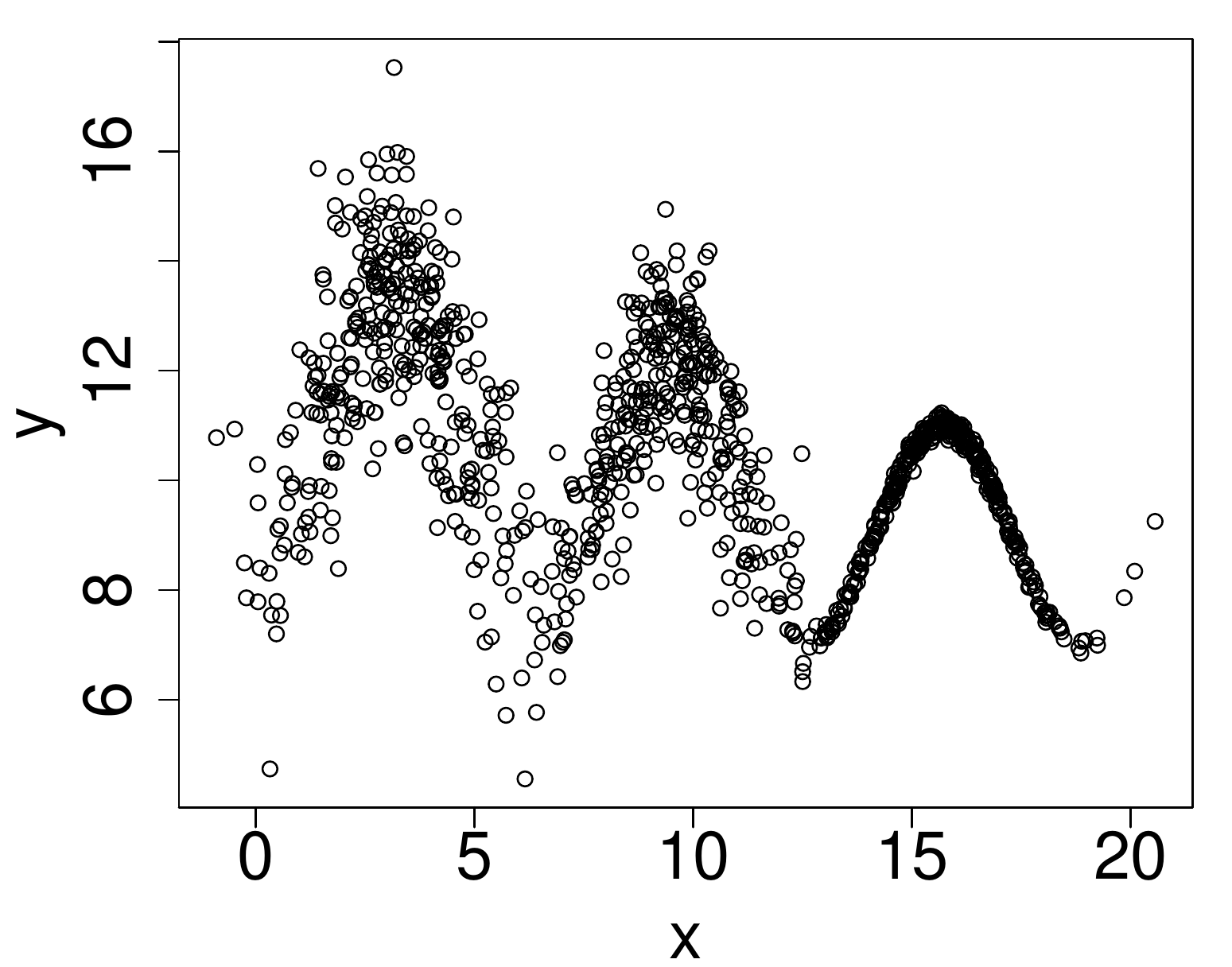} 		
	\end{subfigure}
	\caption{Settings and the generated data for Simulation I-(d). }\label{sim1d-1}
\end{figure}

In this simulation, we discuss the trade-off between the bias and the variance discussed in Section \ref{sec3.2}. The mapping $f(X)$ is 
\begin{equation}\label{sim1d}
    f(X) = \left \{
  \begin{aligned}
    &-2.5(\cos(X)-4.5),  && 0 \leq X < 2\pi \\
    &-2.25(\cos(X)-4.5), && 2\pi \leq X < 4\pi\\
    &-2(\cos(X)-4.5), && 4\pi \leq X \leq 6\pi
  \end{aligned} \right. .
\end{equation}
All the other settings are the same as Simulation I-(c). Function $f(X)$ and the generated data are provided in Fig.~\ref{sim1d-1}. 

 For any candidate $\gamma$ among $\{0, 5, 10, ...,45, 50\}$, the 4-fold cross-validation technique is used to estimate the bias as well as the variance term defined in Section \ref{sec3.2}. As shown by Fig.~\ref{sel2}. With the increase of $\gamma$, the bias term (i.e., the evaluated mean response within the detected region) stays relatively stable at the beginning when the regularization is not large enough to push the region to the latter two sub-regions with lower peaks. When the regularization size reaches a certain level ($\gamma=35$ in this case), the bias increases to its maximal and then stays at that level, as the detected envelope stays within the third sub-regions with smaller mean and less variability. The variance curve displays an inverse trend with the increase of the regularization size. 


\subsection{Simulation study II} \label{sec4.2}
In this section, we consider a simulation study with $p=2$ state variables. We illustrate that it is infeasible to conduct a multidimensional operating envelope identification by separately learning the lower and upper limits on each state variable dimension when the state variables are correlated. We also show the results for different numbers of the disjoint regions. 

\begin{figure}[htbp]
	\centering
	\begin{subfigure}[t]{1.55in}
		\centering
        \caption{Scatter plot of $X_1$ and $X_2$ with distribution $g(X_1, X_2).$}\label{sim2dim-1}
        \vspace{-0.1in}
		\includegraphics[width=35mm]{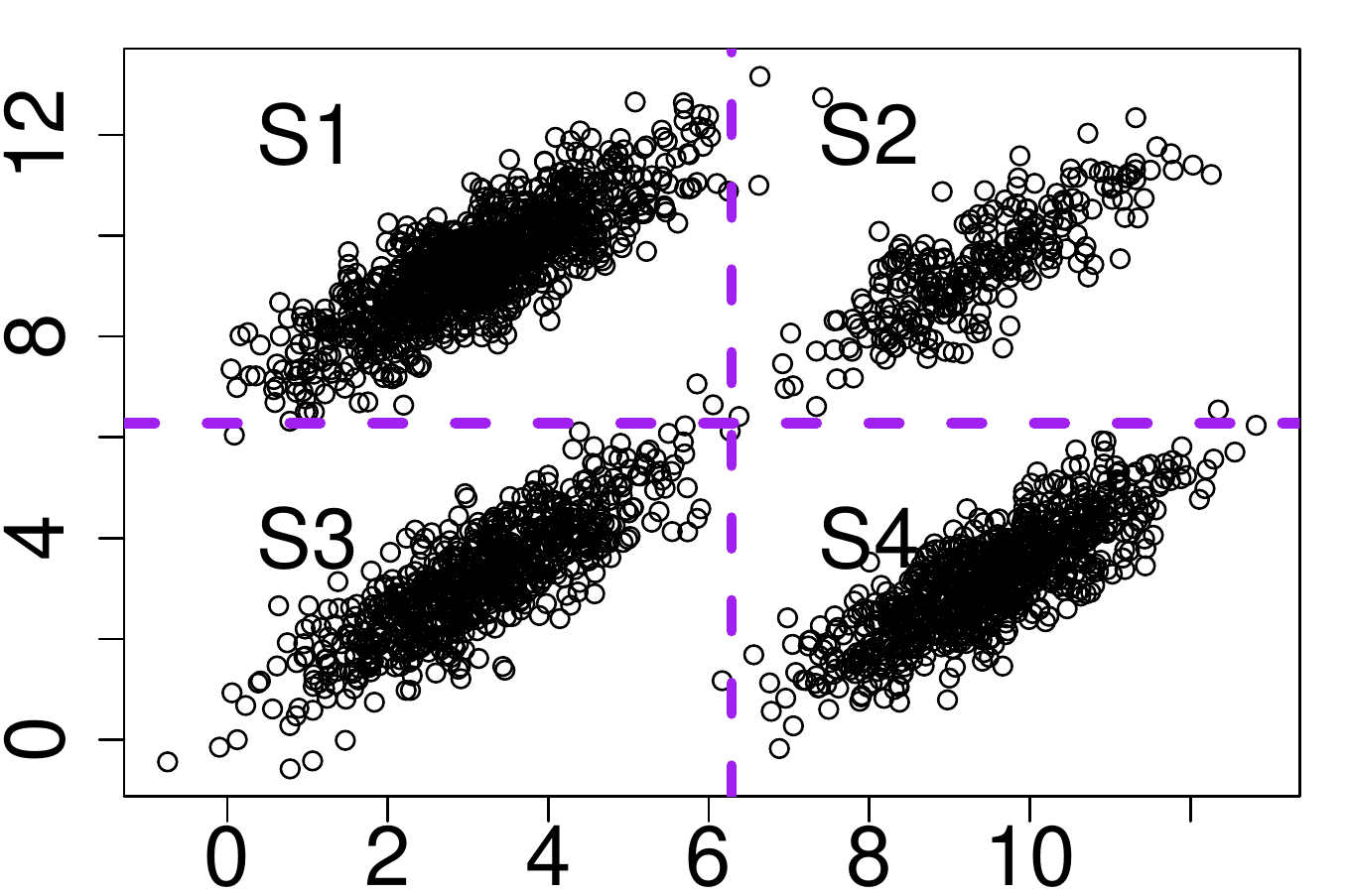} 		
	\end{subfigure}
	\quad
	\begin{subfigure}[t]{1.55in}
		\centering
        \caption{Output using $X_1$ with coverage threshold $\sqrt{\beta}=\sqrt{0.2}$. }\label{sim2dim-2}
        \vspace{-0.1in}
		\includegraphics[width=36mm]{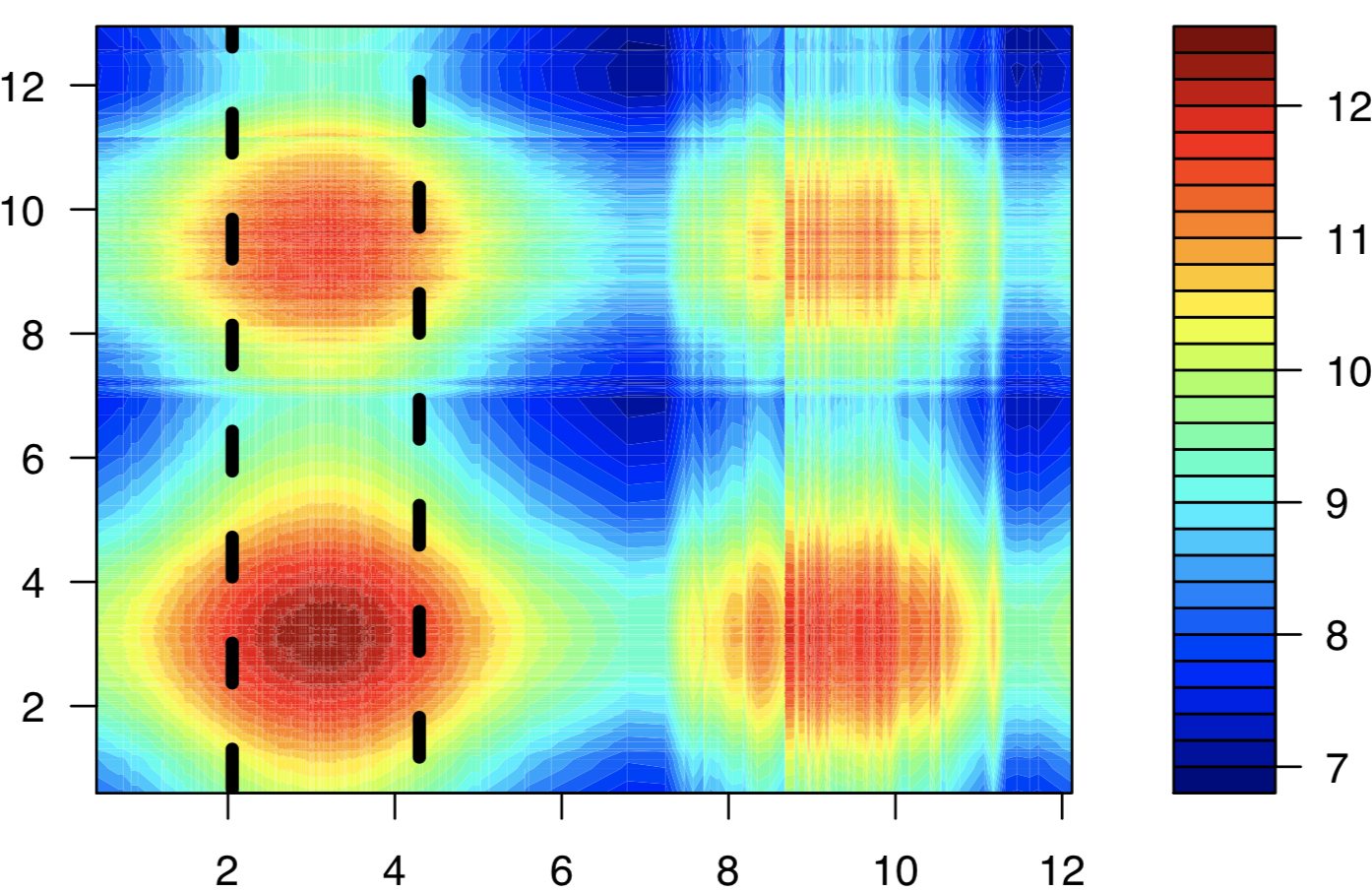} 		
	\end{subfigure}
	\par \bigskip
	\vspace{-0.1in}
	\begin{subfigure}[t]{1.55in}
		\centering
        \caption{Output using $X_2$ with coverage threshold $\sqrt{\beta}=\sqrt{0.2}$. }\label{sim2dim-3}
        \vspace{-0.1in}
		\includegraphics[width=36mm]{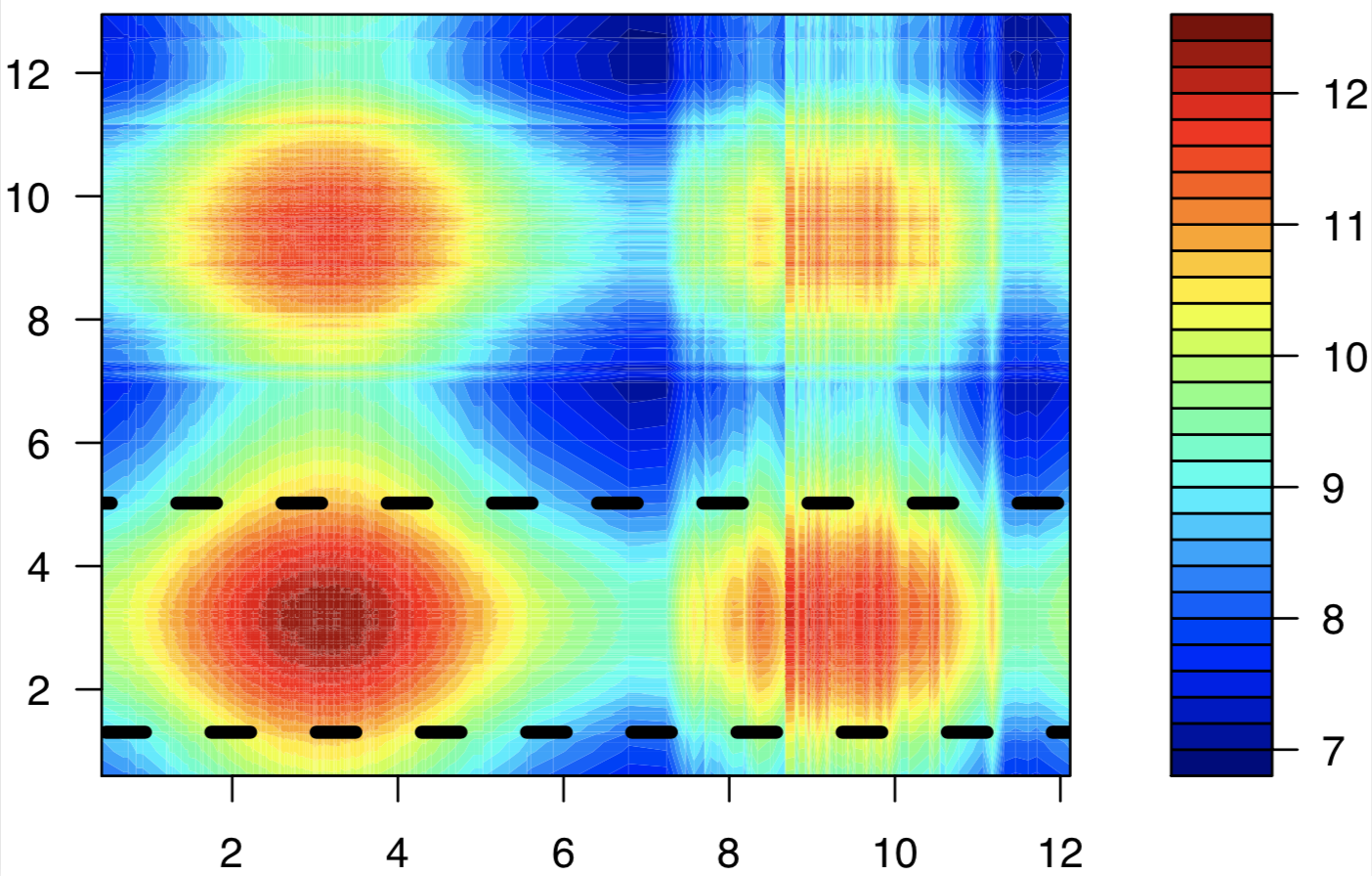} 		
	\end{subfigure}
	\quad
	\begin{subfigure}[t]{1.55in}
		\centering
        \caption{Output using $(X_1,X_2)$, $L=1, \beta=0.2$}\label{sim2dim-4}
        \vspace{-0.1in}
		\includegraphics[width=36mm]{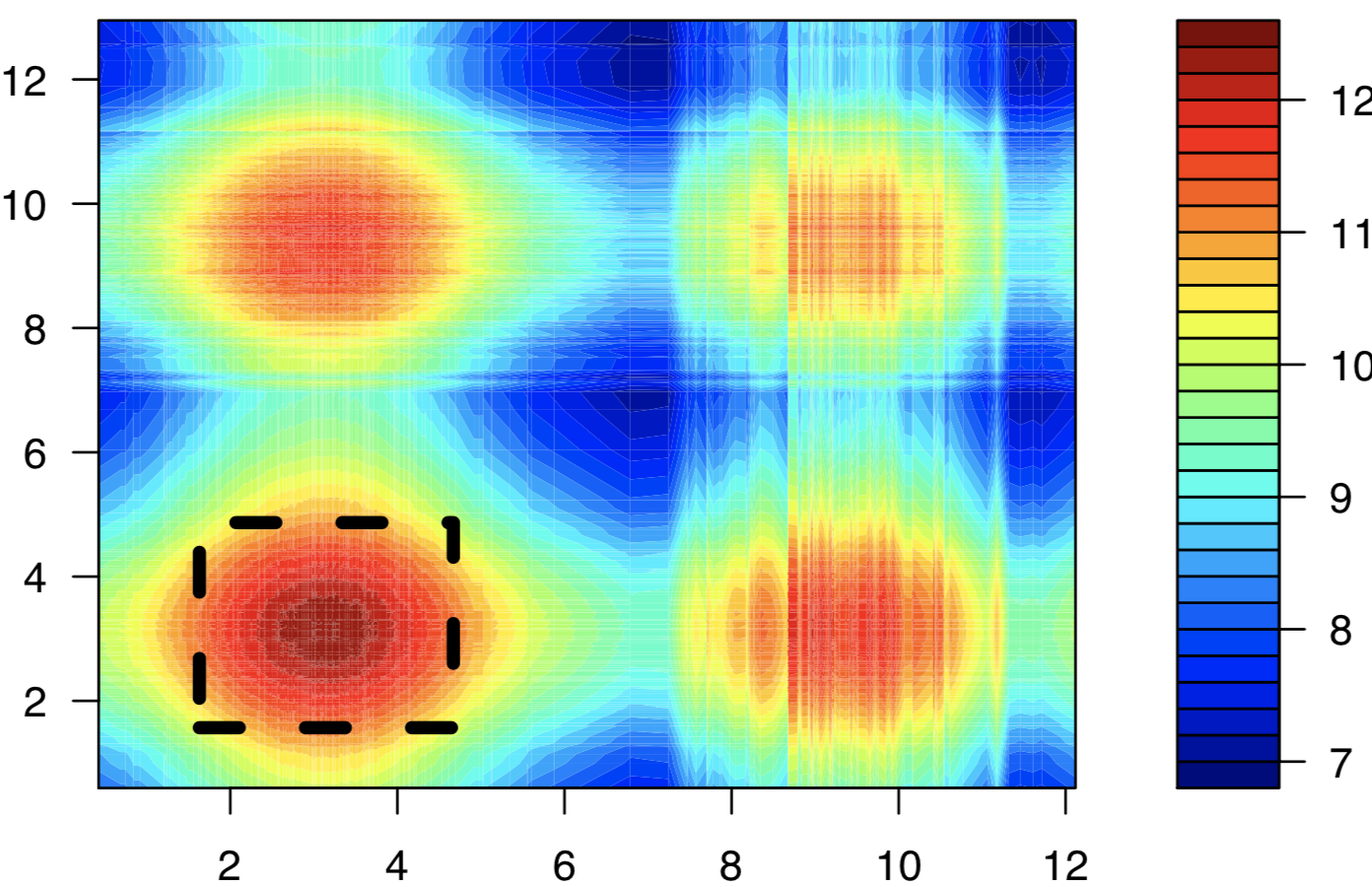} 		
	\end{subfigure}
	\caption{Relevant results in Simulation II when the number of disjoint regions is $L=1$.}\label{sim2dim-1234}
	\vspace{-0.1in}
\end{figure}

The data are simulated as follows. The joint probability density function of the state variable vector is $g(X_1, X_2)=\frac{1}{4}\sum_{s=1}^4N(\ve{\mu}_{s}, \ve{\Sigma}))$, where $\ve{\Sigma} = \left(\begin{smallmatrix} 1.2&1\\ 1&1.2 \end{smallmatrix}\right)$, the means are $\ve{\mu}_{1}=(\pi, \pi)$, $\ve{\mu}_{2}=(\pi, 3\pi)$, $\ve{\mu}_{3}=(3\pi, \pi)$ and $\ve{\mu}_{4}=(3\pi, 3\pi)$. The mapping from $(X_1, X_2)$ to $Y$ is specified as $f(X_1, X_2)=\sqrt{f_0(X_1)f_0(X_2)}$, with $f_0(Z)$ being $-2.25(\cos(Z)-4.5)$ when $0\leq Z \leq 2\pi$ and $-2(\cos(Z)-4.5)$ when $2\pi< Z \leq 4\pi$. The standard deviations of random errors are respectively 0.11, 0.15, 0.05, 0.5 in the four sub-regions ($S_1$, $S_2$, $S_3$, $S_4$) in Fig.~\ref{sim2dim-1}. The contour plot of the generated data can be found in Fig.~\ref{sim2dim-1234}.

Under this simulation setting, it can be seen that we can achieve a better expected region-wise response variable by bounding both state variables. Let's first consider to calculate a single rectangle shaped operating envelope (i.e., $L=1$) under the probabilistic coverage constraint $P(X_1\in [r_{1}^l, r_{1}^u], X_2\in [r_{2}^l, r_{2}^u]) > \beta$. To calculate the 2-dimensional operating envelope, one natural thinking is to bound each state variable separately and then construct a rectangle shaped envelope from the Cartesian product of the detected intervals. However, it is infeasible to appropriately specify the required probabilistic coverage on each state variable dimension, due to the fact that relation in Eq.~\eqref{sim22} no longer holds when the state variables are not independent.
\begin{multline}
\label{sim22}
P(X_1\in [r_{1}^l, r_{1}^u], X_2\in [r_{2}^l, r_{2}^u]) =\\ 
P(X_1\in [r_{1}^l, r_{1}^u]]) P(X_2\in [r_{2}^l r_{2}^u]) 
\end{multline}
In this study, we use simulated data to justify this argument. Specifically, suppose that the required coverage for the targeting rectangle shaped operating envelope is $\beta=0.2$. We calculate the interval on $X_1$ and $X_2$ separately with probabilistic coverage $\sqrt{\beta}=\sqrt{0.2}$. The identified intervals are shown in Fig.~\ref{sim2dim-2} and Fig.~\ref{sim2dim-3}. However, the rectangle determined by the Cartesian product of these two intervals only covers $14.67\%$ of data, which is smaller than the required coverage threshold $\beta=0.2$. A more reasonable approach is to utilize our proposed `GA + penalty' algorithm in Section \ref{sec3} to jointly search over the two-dimensional state variable space. The detected region with a valid empirical coverage ($20.02\% > \beta=0.2$) is visualized in Fig.~\ref{sim2dim-4}. 

\begin{figure}[htbp]
	\centering
	\begin{subfigure}[t]{1.45in}
		\centering
        \caption{$\beta=0.2$}\label{sim2dim-5}
        \vspace{-0.1in}
		\includegraphics[width=36mm, height=23mm]{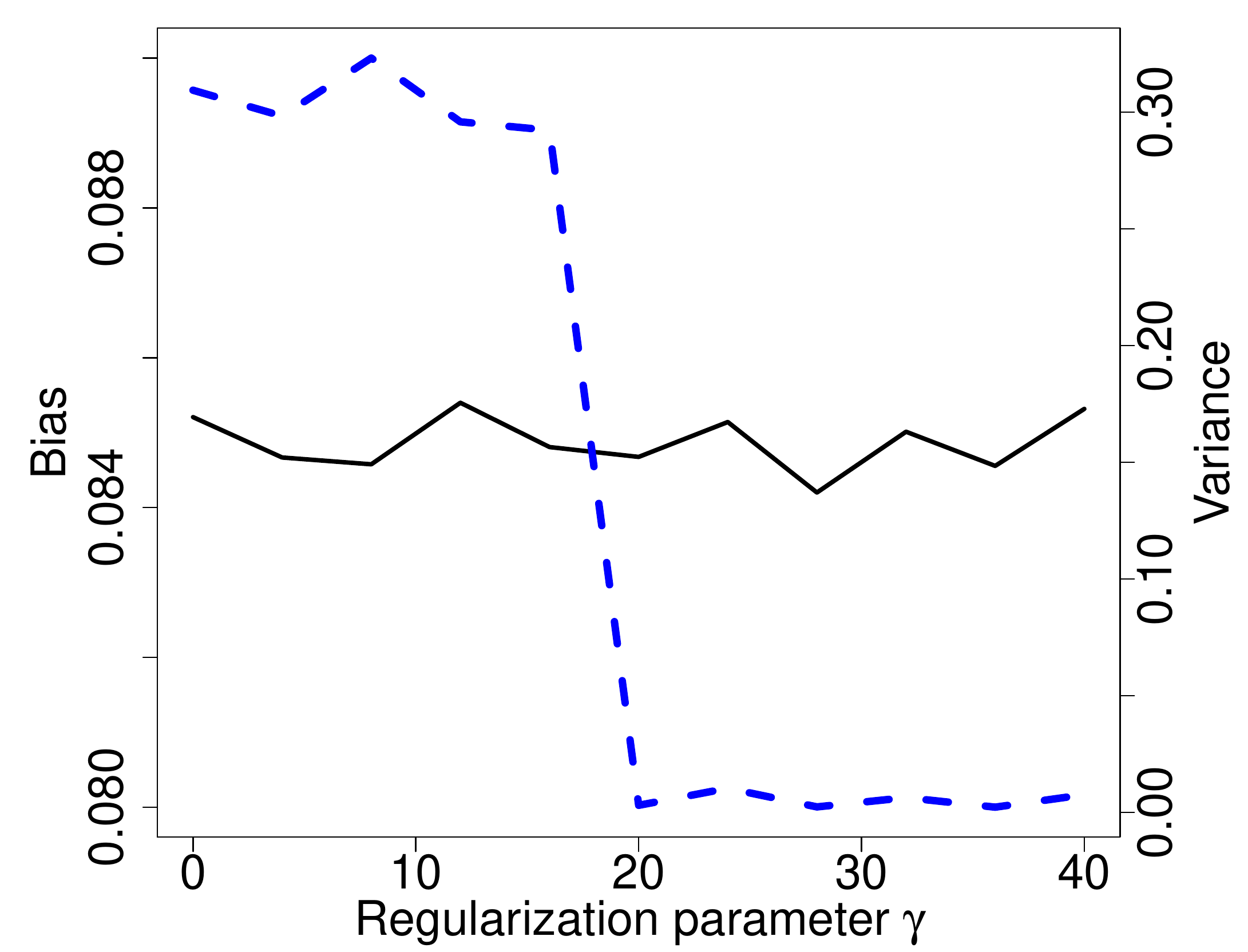} 		
	\end{subfigure}
	\quad
	\begin{subfigure}[t]{1.45in}
		\centering
        \caption{$\beta=0.35$}\label{sim2dim-6}
        \vspace{-0.1in}
		\includegraphics[width=36mm, height=23mm]{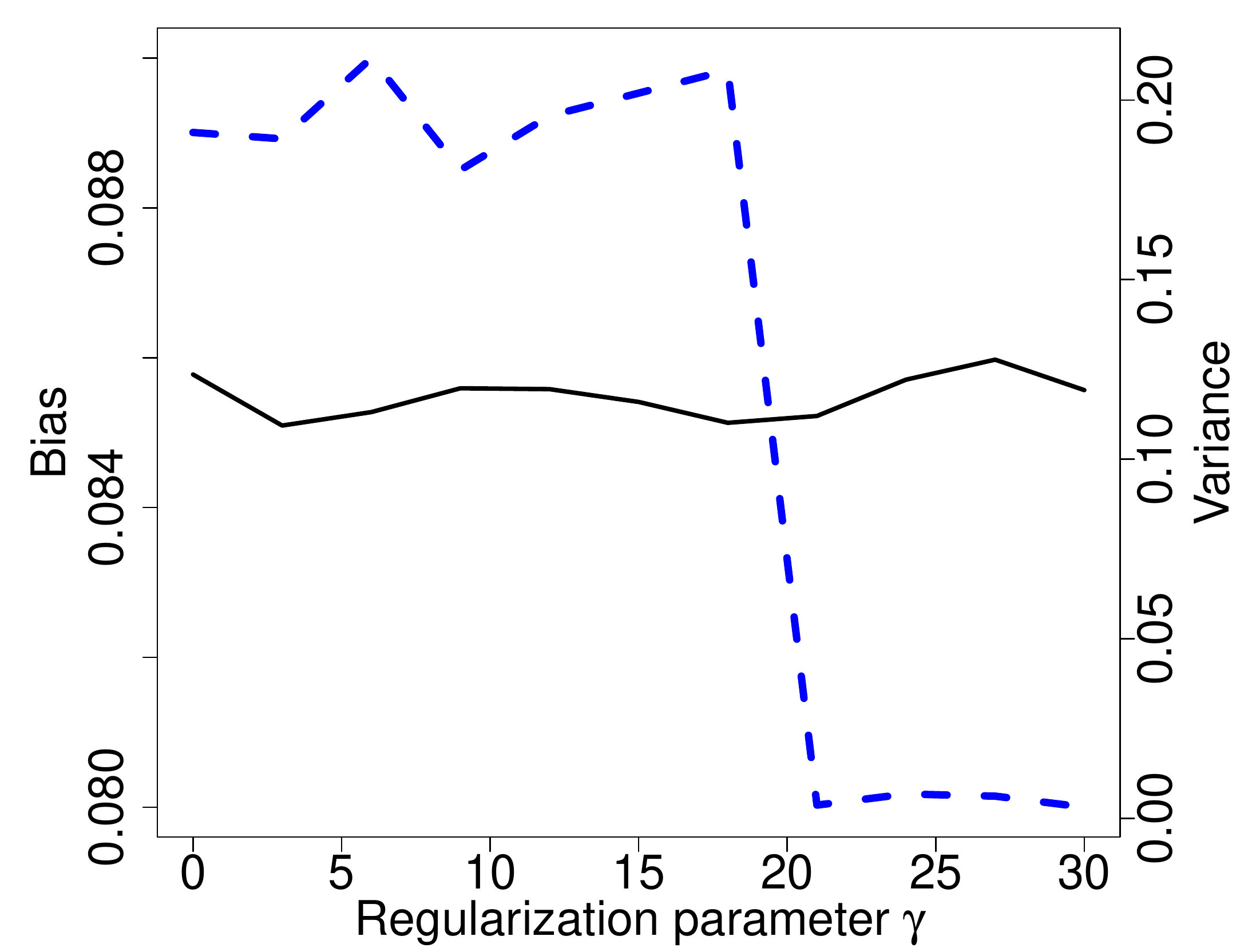} 		
	\end{subfigure}
		\par \bigskip
	\begin{subfigure}[t]{1.45in}
		\centering
        \vspace{-0.1in}
		\includegraphics[width=34mm]{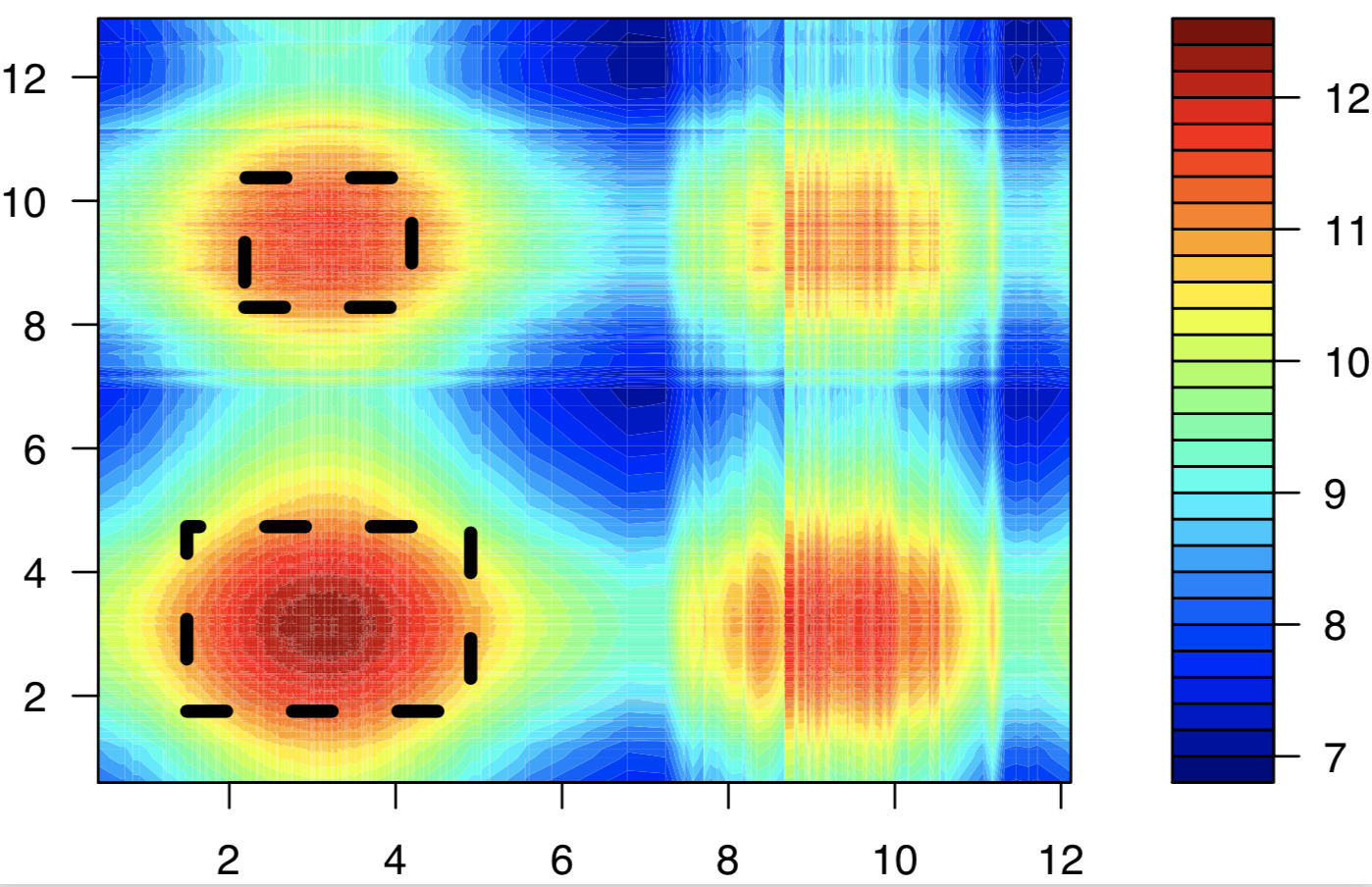} \label{sim2dim-7}		
	\end{subfigure}
	\quad
	\begin{subfigure}[t]{1.45in}
		\centering
        \vspace{-0.1in}
		\includegraphics[width=34mm]{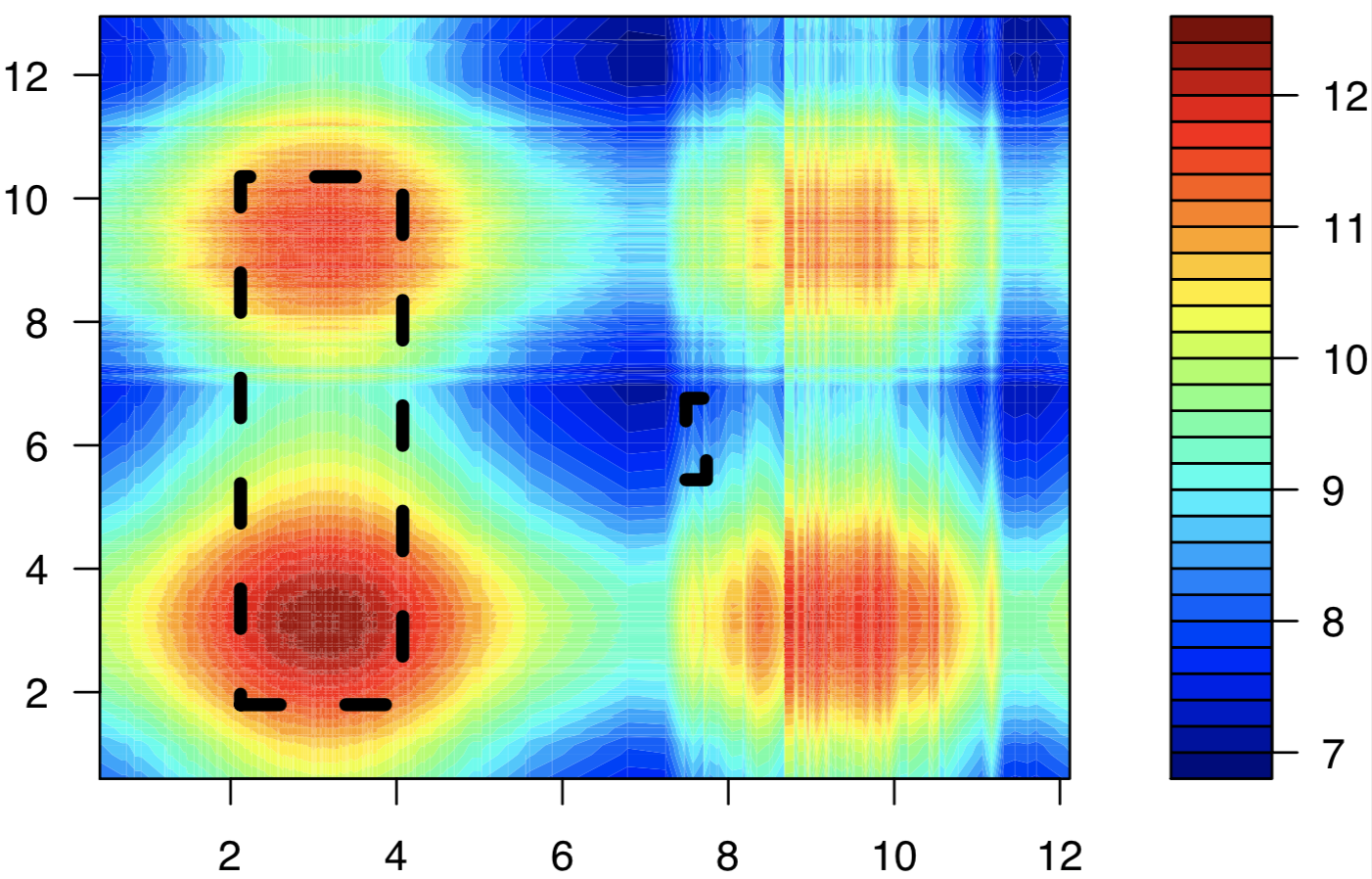} \label{sim2dim-8}		
	\end{subfigure}
	\caption{Relevant results in Simulation II when the number of disjoint regions is $L=2$. In the bias and variance trade-off plots, bias is the black line and variance is the blue line. }\label{sim2dim-5678}
\end{figure}

Next, we evaluate the performance of the proposed regularized `GA + penalty' algorithm when target results consist of $L=2$ disjoint regions. According to our simulation setting, sub-region $S_3$ has the largest mean and the smallest variance (best sub-region). Sub-regions $S_1$ and $S_4$ have the second largest mean values, and the variability within sub-region $S_1$ is smaller than that within sub-region $S_4$. These facts indicate that the detected $L=2$ disjoint envelopes should be within $S_3$ and $S_1$, such that the achieved high mean response variable is more probably to be reached in unseen data. We first set the required probabilistic coverage as $\beta=0.25$, the bias and variance terms as a function of the regularization parameter $\gamma$ are given in Fig.~\ref{sim2dim-5}. Based on these plots, we choose $\gamma=20$ and the resulted envelope is shown at the bottom of Fig.~\ref{sim2dim-5}. The bias and variance curves, and the achieved operating envelope for $\beta=0.35$ are given in Fig.~\ref{sim2dim-6}. The regularization term introduced in Section \ref{sec3.2} helps to push the second envelope into the sub-region $S_1$ with less variability, instead of $S_4$, even though the mean response is the same within these two sub-regions. All the simulation results in Section \ref{sec4.1} and Section \ref{sec4.2} demonstrate the effectiveness of our proposed regularized operating envelope identification algorithm.

\subsection{Operating envelope for mining processes} \label{sec4.3}
In this section, we apply the proposed regularized `GA + penalty' operating envelope identification algorithm to real industrial data on Kaggle.com. The data set is uploaded to Kaggle by Eduardo Magalh\~aes Oliveira and can be downloaded through \url{https://www.kaggle.com/edumagalhaes/quality-prediction-in-a-mining-process}. The data set includes the quality of produced ore in a flotation plant, variables indicating the quality of the raw ore, as well as state variables that quantify the operations along the mining processes. Among all the state variables, the data disposer mentioned that five of them, including starch flow, amina flow, ore pulp flow, ore pulp pH, and ore pulp density, are highly related to the quality of the produced ore. Our objective is to jointly identify an implementable set of upper and lower limits for the five important state variable such that the resulted ore quality is higher or highest given the quality of raw ore fed into the mining processes. Totally, we used 36872 records, and $50\%$ of them are used for training and the remaining $50\%$ data are used for testing. 

First, to move the effect of raw ore's quality on the quality of the produced ore, we propose to normalize the quality of the produced ore by subtracting its fitted value from the random forest regression given the raw ore's quality. The random forest regression is learned from the training data. Next, we apply the proposed `GA + penalty' approach both with and without regularization to identify the region spanned by the top five important state variables such that the corresponding mean normalized quality score of the produced ore is high. The results for coverage threshold $\beta=0.2$, the number of disjoint regions $L=1,2$, and the regularization parameter $\gamma=5, 10$ are summarized in Table \ref{tab1}. As shown by  Table \ref{tab1}, the results match with our intuition. First, when learning without regularization, we achieve a higher optimal mean response when $L=2$ compared to $L=1$. This is because we search over a larger set of candidate solutions when $L$ is higher. Second, given the number of disjoint regions $L$, the bias increases with the increase of regularization parameter, while the variance is smaller for larger regularization parameter values. This phenomenon justifies the bias and variance trade-off discussed in Section \ref{sec3.2}. In practice, users can choose the combinations $L$, $\beta$, $\gamma$ that best meets their needs. 

We notice that the computational cost of our GA based algorithm increases with the increment in dimensions (i.e., the number of state variables considered), due to the nature of GA. For the real data analysis, it took us about 5 hours to get the results. However, as discussed in the previous sections, GA is the most reasonable solution for the considered operating envelope with interpretability and implementability constraints problem. From the application perspective, the operating envelope is not needed to be learned in a streaming mode, hence the computational cost is not a big issue for the proposed `GA + penalty' approach.  



\begin{table}[htbp]
\setlength{\tabcolsep}{2pt}
\caption{Identified operating envelope results on mining processes when $\beta=0.2$. The baseline, i.e., mean quality score over all data points, is $0.0039$. }
\vspace{-0.1in}
\begin{center}
\begin{tabular}{cccccccc}
\hline
\hline
\textbf{L}& \textbf{Method}& &  \textbf{Envelope}&  \textbf{Train}& \textbf{Test} & \textbf{Bias}& \textbf{Var} \\
\hline
$L=1$&Without  &$R=$ &\begin{tabular}{cc}$[2472.63, 4749.30]$\\$[485.73, 726.54]$\\$[388.52, 408.40]$\\$[9.16, 10.22]$\\$[1.53, 1.76]$\end{tabular}& $0.177$ & $0.151$ & 5.65& 0.026\\
\hline
$L=1$& \begin{tabular}{cc}With\\($\gamma=5$)\end{tabular}  &$R=$ &\begin{tabular}{cc}$[2137.81, 4869.11]$\\$[473.28, 672.65]$\\$[393.70, 408.53]$\\$[9.32, 10.04]$\\$[1.52, 1.79]$\end{tabular}& $0.156$  &$0.144$ &  $6.41$ &$0.012$ \\
\hline
$L=1$& \begin{tabular}{cc}With\\($\gamma=10$)\end{tabular}  &$R=$ &\begin{tabular}{cc}$[2121.13, 5923.04]$\\$[474.77, 726.68]$\\$[389.03, 405.60]$\\$[9.15, 10.18]$\\$[1.65, 1.78]$\end{tabular}&  $0.144$ &  $0.135$ &  $6.94$ & $0.009$\\
\hline
$L=2$&Without  &\begin{tabular}{cc}\\ $R_1=$ \\  \\ \\ \\ \\ $R_2=$ \\ \\ \end{tabular}&\begin{tabular}{cc}$[2797.96, 5182.77]$\\$[481.15, 692.58]$\\$[385.51, 408.39]$\\$[9.34, 10.22]$\\$[1.56, 1.82]$\\$[1528.42, 3451.57]$\\$[277.99, 336.60]$\\$[391.10, 416.49]$\\$[9.58, 10.15]$\\$[1.59, 1.79]$\end{tabular}& $0.185$ & $0.163$ & 5.41 & 0.022 \\
\hline
\hline
\end{tabular}
\label{tab1}
\vspace{-0.1in}
\end{center}
\end{table}



\section{Conclusion and Discussion}\label{sec5}
In this paper, we formally categorize the operating envelope problem into three types. For identifying the operating envelope with respect to uncontrollable state variables, we propose a new mathematical definition that not only directly targets the region-wise mean response (the ultimate objective of learning operating envelopes) but also effectively accounts for the interpretability and implementability of the solution. To solve for the new operating envelope with constraints, we propose a regularized `GA + penalty' approach which is capable of calculating the desired interpretable and implementable envelope with the bias and variance terms being well balanced. Our numerical experiments including two sets of simulation studies and the application to a real-world data challenge demonstrate the validity of our proposals.

\balance

\bibliographystyle{IEEEtran}
\bibliography{fanova}

\end{document}